# A Non-Reference Diffusion-Based Restoration Framework for Landsat 7 ETM+ SLC-off Imagery in Antarctica

Leyue Tang, Jonathan Louis Bamber, Gang Qiao, *Member, IEEE*, and Yuanhang Kong

*Abstract*— **Acquiring usable optical imagery in Antarctica is inherently challenging due to prolonged polar nights and frequent cloud cover. Landsat provides the longest and most continuous optical observations and constitutes one of the most important remote sensing data sources for Antarctic studies. However, the scan-line corrector (SLC) failure in 2003 resulted in approximately 22% missing pixels in Landsat 7 ETM+ SLC-off imagery, severely limiting its usability. Unlike many non-polar environments, Antarctic surfaces undergo rapid and substantial changes, which makes it difficult to obtain reliable reference imagery and reduces the applicability of conventional reference-based gap-filling methods. To address this challenge, we propose DiffGF, a non-reference diffusion-based framework for restoring Landsat 7 SLC-off imagery without requiring any external reference data. DiffGF adopts a two-stage design consisting of a latent-space diffusion process and a pixel-space refinement. A dedicated Antarctic dataset, SLCANT, is constructed for training and evaluation. Quantitative and qualitative results demonstrate that DiffGF restores Antarctic SLC-off imagery with high fidelity. Its practical value is further examined through a downstream crevasse segmentation application. The results suggest that DiffGF provides a useful approach for exploiting Landsat 7 SLC-off archives in Antarctica, enabling the extraction of valuable information from historical records and supporting related Antarctic studies.**

*Index Terms*—**Antarctica; diffusion model; image restoration; Landsat 7 SLC-off**

## I. INTRODUCTION

Continuous observation of Antarctica is essential for understanding its mass balance and stability, as well as their implications for future sea-level rise [1, 2]. The Landsat program, operating since 1972, has provided the longest continuous earth observation record worldwide [3]. Owing to its unparalleled temporal continuity, Landsat imagery has been one of the most important optical remote sensing data sources for Antarctic research, and has profoundly impacted a wide range of Antarctic studies [4-7]. Among the Landsat series, Landsat 7 and its Enhanced Thematic Mapper Plus (ETM+) play a key role in global monitoring. However, the permanent failure of its scan-line corrector (SLC) in May 2003 resulted in approximately 22% of pixels remaining unscanned (hereafter referred to as missing or invalid pixels) [8]. The acquired imagery, commonly referred to as SLC-off imagery, suffered from data gaps, severely limiting the usability of subsequent observations. Prior to the launch of Landsat 8 in 2013, this decade of SLC-off imagery constituted one of the most critical and irreplaceable optical data sources for Antarctica. Consequently, the data quality issues associated with SLC-off imagery substantially constrained Antarctic research during this period, particularly time-series studies that rely on reliable and temporally continuous remote sensing observations.

After the SLC failure, the United States Geological Survey (USGS) proposed reconstructing SLC-off imagery by using valid scanlines and auxiliary reference data [8]. Since then, various restoration methods have been developed, which can generally be categorized into two groups: (1) non-reference methods that rely solely on the SLC-off imagery itself, and (2) reference-based methods that incorporate additional reference.

Non-reference methods reconstruct missing pixels using only the valid information within the SLC-off image itself, without requiring external data. Conventional non-reference approaches typically assume statistical or structural continuity between known and unknown regions [9]. Classical methods, such as the widely used Gapfill tool in ENVI and kriging-interpolation-based method [10], estimate missing pixels from adjacent valid pixels. Although these methods can produce complete imagery, their ability to effectively reconstruct structural and semantic details remains limited, particularly in heterogenous or complex landscapes [11, 12].

Reference-based methods restore missing areas by leveraging external auxiliary data, such as multi-temporal observations or data from other sensors. One of the earliest and most representative approaches is the local linear histogram matching (LLHM) method proposed by USGS shortly after the SLC failure, which estimates missing pixel values by applying a local linear transformation derived from corresponding pixels in multi-temporal reference image(s) [8]. However, this approach is sensitive to radiometric differences

This work is supported by the National Natural Science Foundation of China (42276249, 42394131), the Science and Technology Commission of Shanghai Municipality (23230712200), and the Fundamental Research Funds for the Central Universities. (Corresponding author: Gang Qiao).

Leyue Tang is with Center for Spatial Information Science and Sustainable Development Applications, College of Surveying and Geo-Informatics, Tongji University, Shanghai 200092, China, and Bristol Glaciology Centre, School of Geographical Sciences, University of Bristol, Bristol, UK (email: tly22@tongji.edu.cn and leyue.tang@bristol.ac.uk).

Jonathan Louis Bamber is with Bristol Glaciology Centre, School of Geographical Sciences, University of Bristol, Bristol, UK, and Institute for Advanced Study, Technical University of Munich, Germany (email: J.Bamber@bristol.ac.uk).

Gang Qiao is with Center for Spatial Information Science and Sustainable Development Applications, College of Surveying and Geo-Informatics, Tongji University, Shanghai 200092, China (e-mail: qiaogang@tongji.edu.cn).

Yuanhang Kong is with College of Surveying and Geo-Informatics, Tongji University, Shanghai 200092, China (email: 2111330@tongji.edu.cn).

caused by cloud, illumination, or land surface dynamics. Later advancements introduced more sophisticated modeling strategies to improve reconstruction quality and robustness. Neighborhood Similar Pixel Interpolator (NSPI) [13] predicts missing values using spectrally and spatially similar pixels, geostatistical neighborhood similar pixel interpolator (GNSPI) [14] enhances this by incorporating geostatistical modeling. Weighted linear regression with Laplacian prior regularization (WLR-LPRM) [15] fills the gaps by building a regression model. Spatial–spectral radial basis function (SSRBF)-based interpolation method [16] performs global linear histogram matching followed by spatial-spectral weighted interpolation. More recently, progressive gap-filling through the cascading temporal and spatial framework (PGFCTS) [17] employs a cascading manner to sequentially integrate temporal texture reconstruction and spatial spectral refinement.

While reference-based approaches generally outperform classical interpolation-only methods in reconstruction quality, their practicality and effectiveness heavily depend on the availability and quality of suitable reference data, as well as stability of the surface properties. Variations due to cloud cover and surface dynamics may limit their applicability, especially in scenarios with scarce or outdated auxiliary reference observations.

Antarctica represents a particularly challenging environment for reference-based SLC-off restoration, where the fundamental assumptions of reference-based approaches may not always hold in practice, reducing their reliability and practical applicability. The prolonged polar night severely restricts the temporal window for acquiring optical imagery, and frequent cloud cover further reduces the availability of cloud-free imagery. More importantly, even when cloud-free images are available, the Antarctic surface, especially over ice shelves, can undergo much more rapid and substantial changes than most non-polar environments. As a result, available images may not always provide physically consistent reference information, posing a distinct challenge compared with many other scenarios. Therefore, the development of non-reference restoration methods for SLC-off imagery is particularly valuable for Antarctic remote sensing applications.

In recent years, deep learning (DL) has gained increasing attention in image inpainting tasks. Since the first application of convolutional neural networks (CNNs) to image inpainting [18], a wide range of DL-based methods have been developed [19-21]. Among them, generative adversarial network (GAN)-based methods have shown strong performance, such as LaMa [22] that once represented the state-of-the-art (SOTA) method.

In the meantime, DL-based approaches have also been applied to SLC-off imagery reconstruction. A CNN-based network utilizing spatial-temporal-spectral auxiliary inputs was proposed in [23] to reconstruct the missing data in remote sensing imagery, with Landsat 7 SLC-off restoration demonstrated as a representative application, alongside Aqua MODIS band 6 dead-line reconstruction and thick cloud removal. Two other DL-based reconstruction frameworks, based on radiation decoupling [24] and local-global harmonization [25], have also demonstrated applicability to SLC-off restoration using auxiliary reference images. More recently, an edge-conditional Transformer-based inpainting method was proposed in [26], which uses edge information to guide attention toward high-frequency details and critical geomorphological structures, demonstrating its effectiveness for SLC-off gap filling on non-polar imagery. However, its applicability to polar environments remains to be further investigated. Non-reference efforts include the application of Deep Image Prior (DIP) [27], which optimizes an image-specific CNN directly on the known pixels in the SLC-off imagery. In addition, a GAN-based model has been proposed to target the reconstruction of the panchromatic band [28]. Nevertheless, these methods exhibit limitations in broader applications. DIP, though effective without pretraining or external data, is sensitive to parameter settings and may require multiple runs to achieve optimal results. The GAN-based method targets only a single band, making it less suitable for scenarios that demand richer spectral information. A residual deep neural network (DNN) designed for n-line striping and noise removal has also been tested for SLC-off restoration [29]. However, as this model was trained for vertical stripe noise removal, its application to oblique SLC-off line loss requires additional preprocessing, which may constrain its applicability to general SLC-off restoration scenarios.

Despite these advances, DL-based methods tailored specifically for non-reference SLC-off restoration remain relatively underexplored. Given their capacity to model complex patterns without auxiliary data, deep learning techniques hold potential for single-image-based SLC-off restoration. This highlights the need for frameworks capable of reconstructing complex spatial structures and preserving semantic content without relying on external reference, which is particularly critical for Antarctic applications.

Diffusion models have recently emerged as a leading paradigm in image synthesis due to their strong ability to generate high-fidelity images with stable training dynamics, effectively avoiding issues such as mode collapse commonly observed in GAN-based approaches [30]. A representative foundational work is the Denoising Diffusion Probabilistic Model (DDPM) [31], which consists of a forward process that incrementally adds noise to the input data to corrupt it via a predefined Markov chain, and a learned reverse process that restores the image from noise. Most subsequent diffusion-based models adopt similar forms.

In the context of image inpainting, diffusion models have demonstrated strong potential. RePaint [32], one representative work, employs an unconditionally trained DDPM and conditions the reverse process with the mask to restore missing regions. However, pixel-space diffusion models like RePaint are computationally expensive due to high-dimensional processing. To improve efficiency, latent-space diffusion models such as Latent Diffusion Model (LDM) [33] compress the image into a latent space representation before the diffusion process. Recent studies have further demonstrated the effectiveness of diffusion models in image inpainting, highlighting their capacity to reconstruct missing image content [34-37].

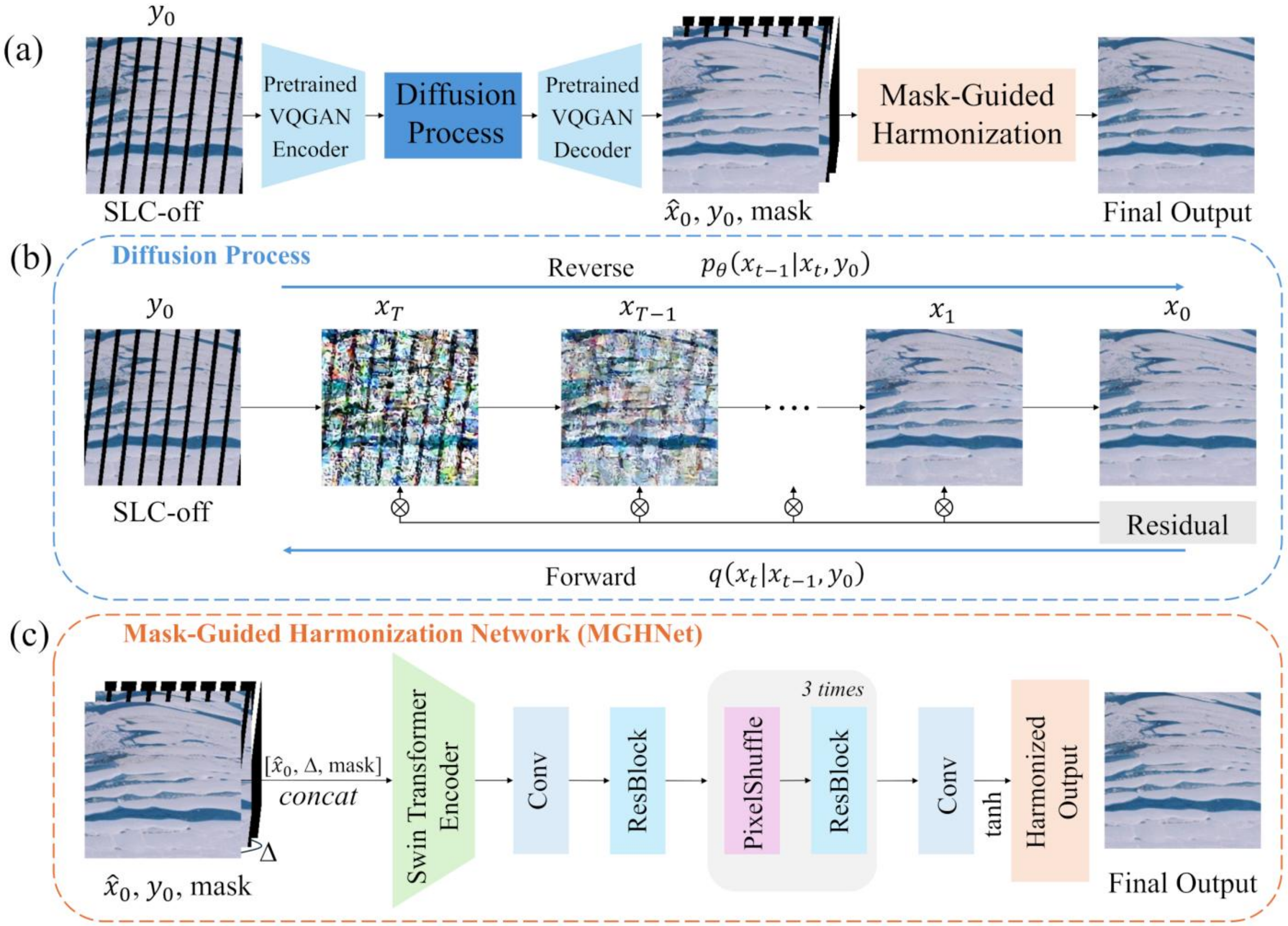

**Fig. 1.** Workflow of the proposed DiffGF framework. (a) Overview of DiffGF framework, consisting of two stages: a diffusion process followed by a refinement stage using a mask-guided harmonization network for final output. (b) Illustration of the forward and reverse processes in the diffusion model. Intermediate results are decoded into pixel space for intuitive visual interpretation. (c) Structure of the proposed Mask-Guided Harmonization Network (MGHNet), designed to harmonize the reconstructed and known regions.

Beyond inpainting, diffusion models have been successfully applied to a wide range of tasks, including image super-resolution, style transfer, and natural language processing. Diffusion models have recently been extended to remote sensing tasks, including cloud removal [38-41], super resolution [42-45], and image generation [46-48], demonstrating their ability to model complex spatial and spectral patterns in remote sensing imagery. However, the SLC-off problem differs from the above tasks, as it is characterized by systematic, parallel, stripe-shaped data gaps rather than spatially continuous data missing or image-wide enhancement. In Antarctica, this problem is further challenged by the scarcity of suitable auxiliary reference imagery, the relatively weak textural features over snow and ice surfaces, and the need to preserve subtle surface structures across scan-line gaps. In addition, practical deployment in remote sensing applications still faces challenges, particularly in terms of computational cost and inference time under resource-constrained environments [49]. These are important considerations given the huge volume of Landsat 7 data that needs to be corrected.

Motivated by these developments, we explore the potential of diffusion models for Landsat 7 ETM+ SLC-off imagery restoration in Antarctica. To the best of our knowledge, this work is the first to apply a diffusion model in this task and also the first dedicated to Antarctic imagery. Unlike reference-based methods that require temporal or cross-sensor data, our method operates solely on a single SLC-off image, without requiring any external reference data. This makes it especially useful for scenarios where supplementary data is unavailable or outdated, such as Antarctica, with limited alternative observations. In this case, existing SLC-off imagery remains a critical but underutilized resource. Our objective is to reconstruct these images efficiently and effectively using the generative capability of diffusion models, thereby facilitating further downstream analysis of Antarctic imagery.

In this work, we propose DiffGF, a diffusion-based framework designed to restore Landsat 7 ETM+ SLC-off imagery. The framework follows a two-stage design: the first stage performs a diffusion process in the latent space for efficient training and inference while the second stage refines the results in the pixel space to enhance spatial and spectral consistency. A dataset dedicated to Antarctica comprising 3,139 images (256×256 pixels) is constructed, covering diverse surface characteristics across different regions. We further examine the utility of the reconstructed images in a downstream crevasse segmentation task, providing evidence of the potential applicability of DiffGF. The framework is also compared against three representative non-reference methods from both SLC-off and general image inpainting domains.

The main advances of this work are as follows:

1) A diffusion-based framework (DiffGF) is proposed for non-reference SLC-off imagery restoration, performing latent-space diffusion for efficient restoration and pixel-space refinement for harmonization.

2) A Mask-Guided Harmonization Network (MGHNet) is designed to mitigate the information loss introduced by the transformation between latent and pixel spaces, enhancing structural and spectral consistency between reconstructed and known regions.

3) A dedicated Antarctic dataset is constructed, encompassing diverse surface characteristics across different regions.

4) The framework shows promising performance across different Antarctic regions and in a downstream crevasse segmentation application.

## II. Methodology

### *A. DiffGF Framework*

The proposed DiffGF framework addresses the gap-filling challenge in Landsat 7 ETM+ SLC-off imagery through a two-stage process: latent-space diffusion-based initial reconstruction followed by pixel-space refinement using mask-guided harmonization (Fig. 1). Specifically, the SLC-off image is first processed by a diffusion model to initially reconstruct missing regions, and subsequently refined by the Mask-Guided Harmonization Network (MGHNet) to enhance consistency between the reconstructed and known regions. The components of the framework are detailed in the following subsections.

### *B. Diffusion Model*

The initial restoration stage is performed via a diffusion process that reconstructs missing regions based on the original SLC-off image. To improve efficiency and reduce computational costs, the diffusion process is conducted in latent space, where the input image is first encoded into a compact latent representation using a pretrained autoencoder. Unlike conventional diffusion models that initialize from pure Gaussian noise, the reverse process adopted in this framework starts directly from the SLC-off image and is guided by a residual-based transition mechanism [50]. This strategy enables high-quality reconstruction within a few steps while preserving fine-grained image details.

Let $x_0$ and $y_0$ represent the complete and corresponding SLC-off images, respectively. The forward process progressively introduces the residual information $e_0 = y_0 - x_0$ to transform $x_0$ into a degraded version that approximates $y_0$. In the forward process, the variable $x_t$ at timestep $t$ can be computed from $x_0$, $y_0$, and timestep $t$ according to the predefined noise schedule as:

$$q(x_t|x_0, y_0) = \mathcal{N}(x_t;\ x_0 + \eta_t e_0, \kappa^2\eta_t \boldsymbol{I}), t\ =\ 1,\cdots,T \quad (1)$$

where $\eta_t$ is determined by the noise schedule [50], $\kappa$ is a hyperparameter deciding the noise level, $\boldsymbol{I}$ denotes the identity matrix.

The reverse process aims to estimate the posterior distribution of $x_{t-1}$, given $x_t$, $y_0$, and timestep $t$ as:

$$p_\theta(x_{t-1}|x_t, y_0) =\ \mathcal{N}(x_{t-1};\ \mu_\theta, \Sigma_\theta) \quad (2)$$

where the mean parameter $\mu_\theta$ and variance parameter $\Sigma_\theta$ are functions of $x_t$, $y_0$, and timestep $t$. Specifically, the mean includes a weighted combination of $x_t$ and the output of $f_\theta(\cdot)$, which is trained to predict the high-quality complete image $x_0$ with a learnable parameter $\theta$. It is computed as $\mu_\theta = \frac{\eta_{t-1}}{\eta_t} x_t + \frac{\alpha_t}{\eta_t} f_\theta(x_t, y_0, t)$, where $\alpha_1 =\ \eta_1,\ \alpha_t = \eta_t - \eta_{t-1}\ (t > 1)$. The variance is computed as $\Sigma_\theta(x_t, y_0, t) =\ \kappa^2 \cdot \frac{\eta_{t-1}}{\eta_t} \cdot \alpha_t \cdot \boldsymbol{I}$.

During training, $x_t$ at any timestep t ∈ [0,T] is obtained from the forward process defined in Eq. (1), given $x_0$, $y_0$ and the predefined noise schedule. The neural network $f_\theta(\cdot)$ predicts $\hat{x}_0^t$ based on $x_t$, $y_0$, and t, and is trained by minimizing a composite loss function that combines an L2 loss in latent space with a perceptual loss computed in pixel space [50]:

$$\mathcal{L}_\theta(x_t, y_0, t) = \sum_t \|\hat{z}_0^t - z_0\|_2^2 + \lambda \mathcal{L}_p(\hat{x}_0^t, x_0) \quad (3)$$

where $\hat{z}_0^t$ and $z_0$ denote the latent representations of the predicted and ground-truth complete images, respectively, while $\hat{x}_0^t$ and $x_0$ denote their corresponding pixel-space representations. $\lambda$ is a weighting hyperparameter balancing the two terms, and $\mathcal{L}_p(\cdot,\cdot)$ denotes the perceptual loss computed with the pretrained Learned Perceptual Image Patch Similarity (LPIPS) metric [51].

During inference, the original SLC-off image $y_0$ is perturbed with Gaussian noise to generate the $x_T$. Given $x_T$, the timestep $T$, and $y_0$, the neural network $f_\theta(\cdot)$ predicts the $\hat{x}_0^T$, which is then used in the reverse process $p_\theta(x_{t-1}|x_t, y_0)$ to iteratively derive $x_{T-1}, x_{T-2}, \ldots, x_0$.

It is important to note that the entire diffusion process is performed in latent space, although the variables $y_0$, $x_t$, and $x_0$ are presented using their pixel-space notations for simplicity. To avoid introducing excessive symbols, variable names are not explicitly distinguished between latent and pixel spaces, except where necessary (e.g., in the loss function Eq. (3), where both spaces are involved).

The denoising U-Net is an important component in diffusion architectures, contributing significantly to reconstruction quality. In this framework, a modified U-Net [50] is adopted as the denoising network, in which the standard self-attention layers are replaced with Swin Transformer [52]. Benefiting from the local window-based attention mechanism of the Swin Transformer, the network demonstrates improved adaptability to varying image resolutions while maintaining reconstruction performance. Although image resolution is beyond the scope of this study, this design facilitates broader applicability to diverse datasets and application scenarios.

The transformation between pixel and latent spaces is achieved via a pretrained autoencoder. Specifically, a pretrained VQGAN [53] is employed, as its adversarial and perceptual training objectives can help reduce blurring and preserve structural details during image reconstruction from latent representations. Input images are encoded into latent representations for the subsequent diffusion process, substantially reducing the spatial resolution and enabling more efficient and stable diffusion modeling. After the diffusion process is completed, the output is decoded by the VQGAN decoder to generate the reconstruction in pixel space.

### *C. Refinement stage: Mask-Guided Harmonization Network*

Performing the diffusion process in the latent space significantly improves the efficiency of the model and reduces computational requirements, thereby enhancing its practical applicability. However, the information loss introduced by the transformation between latent and pixel spaces remains an inherent limitation of most autoencoders [54]. This issue is particularly pronounced in the SLC-off restoration task due to the spatially discontinuous nature of stripe-shaped missing regions. As a result, visual discontinuities tend to appear at the boundaries between the reconstructed and the surrounding known regions, leading to noticeable artifacts in the reconstructed complete image. These discontinuities not only degrade overall visual quality but also compromise semantic interpretability in downstream applications.

To address the inconsistency between reconstructed and known regions, we design the Mask-Guided Harmonization Network (MGHNet) (Fig. 1(c)) as the refinement stage. The goal of MGHNet is to refine the reconstructed regions by predicting pixel-level corrections that improve their spatial and spectral consistency with the surrounding known regions.

MGHNet adopts an encoder-decoder architecture, where the encoder is implemented using a Swin Transformer encoder [52], for its strong ability in capturing semantic and spatial information through hierarchical representations and shifted window attention. The network is built on the assumption that the differences between the predicted image and the ground truth in masked (missing) and unmasked (known) regions should be spatially and semantically correlated. This implies that variation patterns in the unmasked regions can provide informative cues for estimating the differences in masked regions.

Based on this assumption, the network takes the predicted image generated by the diffusion process ($\hat{x}_0$), the SLC-off image $y_0$, and the binary mask of the missing pixels ($m$) as inputs. The mask serves as explicit guidance, enabling the network to focus on the transition boundaries between reconstructed and known regions, thereby improving visual coherence.

Specifically, the pixel-wise difference of unmasked regions $\Delta = (y_0 - \hat{x}_0) \times (1 - m)$ is first computed. Then, $\hat{x}_0$, $\Delta$, and $m$ are concatenated to form a 7-channel input fed into the encoder. The decoder consists of three components:

(1) a convolutional layer followed by a ResBlock (Fig. 2(b));

(2) three upsampling stages, each composed of a PixelShuffle block (Fig. 2(a)) [55] and a ResBlock (Fig. 2(b));

(3) a final convolutional layer followed by a hyperbolic tangent (tanh) activation, which generates the correction output.

The predicted correction (denoted as $\hat{r}$) is added to $\hat{x}_0$, and the corrected result is used to replace the masked regions in $y_0$, yielding the final reconstruction $\hat{I}$, while strictly preserving the original known pixels, as shown in Eq. (4):

$$\hat{I} = (\hat{x}_0 + \hat{r}) \times m + y_0 \times (1 - m) \tag{4}$$

where m is a binary mask indicating missing pixels (1 for missing pixels, and 0 for known pixels).

The ResBlock used in MGHNet (Fig. 2(b)) is customized from the classical design [56] by removing the Batch Normalization (BN) layer, as BN may disrupt local pixel distributions in this pixel-level correction task. Additionally, all activation functions in the decoder are set to tanh to constrain the output range and preserve fine-grained detail. This configuration is empirically found to yield stable and effective results in this task.

The training of MGHNet is guided by a composite loss function that combines a foreground-normalized L2 loss over the missing regions (adapted from [57]) and a perceptual loss over the complete image. This formulation promotes fine-grained pixel-level accuracy in the missing pixels while ensuring the overall perceptual quality of the harmonized output:

$$\mathcal{L}_{MGHNet} = \lambda \cdot \frac{\|\hat{I} \odot m - I \odot m\|_2^2}{\max(A_{min}, \sum m)} + \mathcal{L}_p(\hat{I}, I) \tag{5}$$

where $\hat{I}$ and $I$ denote the harmonized and ground-truth images, respectively, $\odot$ denotes element-wise multiplication, $\lambda$ is a weighting hyperparameter balancing the two terms, $A_{min}$ is a hyperparameter stabilizing training in the presence of extremely small missing regions, and $\mathcal{L}_p(\cdot,\cdot)$ refers to the perceptual loss computed using the pretrained LPIPS metric.

During inference, MGHNet takes the complete image $\hat{x}_0$ predicted by the diffusion process, along with the original SLC-off image $y_0$ and the corresponding mask, to generate the final output. This refinement step mitigates visual discontinuities between reconstructed and known regions, thereby improving overall visual quality and enhancing reliability of downstream applications.

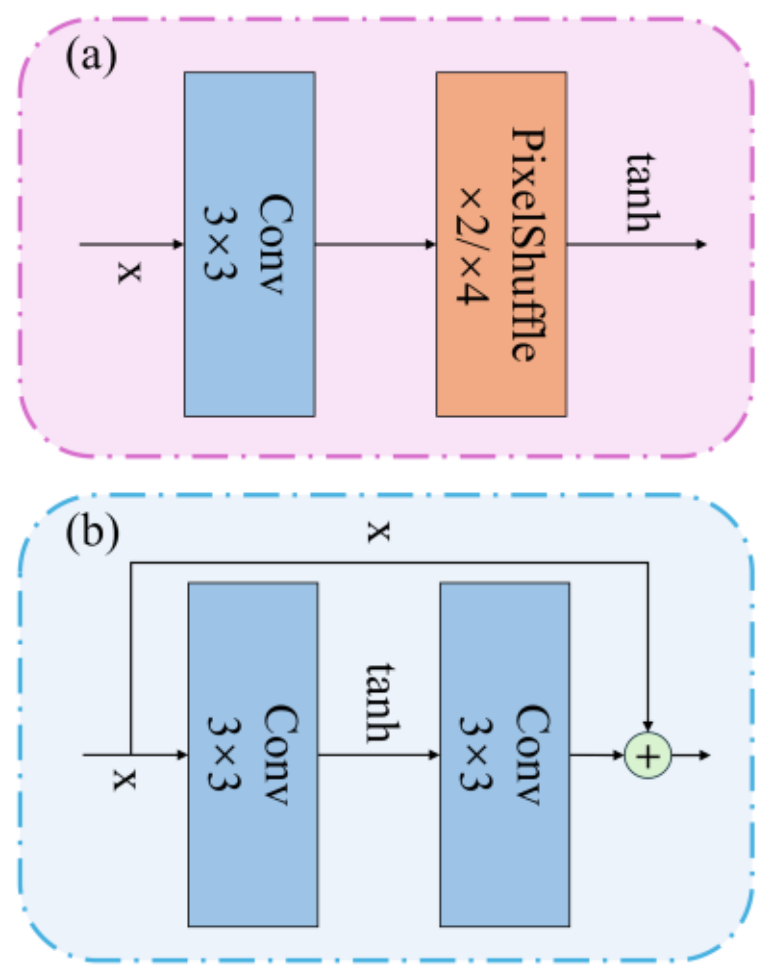

**Fig. 2.** Architectural details of the PixelShuffle Block (a) and ResBlock (b) used in MGHNet.

*D. Training Strategy*

The training of the DiffGF framework is carried out in two separate stages: the denoising network in the diffusion process and the refinement MGHNet are trained independently. The denoising network is trained using the simulated SLC-off images, their corresponding ground truth images, and associated masks, as described in Section III-A.

Although MGHNet is designed to refine the reconstructed image generated by the diffusion model with the original SLC-off image, it is not trained directly on the outputs of the diffusion model. This is to avoid introducing additional variations stemming from the stochasticity of the diffusion process, which may interfere with the objective of correcting artifacts caused by latent–pixel space transformation.

To decouple this effect and effectively learn the harmonization task, we simulate information loss by encoding and decoding the complete ground truth images using the pretrained VQGAN autoencoder, as described in Section III-B. This strategy introduces pixel-level mismatches without the added complexity of generative variation, thereby enabling MGHNet to focus specifically on harmonizing the reconstructed regions with the known regions.

Experimental results demonstrate that MGHNet trained in this manner generalizes well to actual diffusion outputs, effectively refining the predicted images and producing visually consistent reconstructions.

Once trained, DiffGF restores SLC-off imagery using only the input image and its gap mask, without requiring reference imagery during inference.

*E. Evaluation Metrics*

To evaluate the performance of the proposed DiffGF framework in this Landsat 7 ETM+ SLC-off restoration task, we adopt several widely used metrics: Peak Signal-to-Noise-Ratio (PSNR) [58], Universal Image Quality Index (UIQI) [59], Structural Similarity Index Measure (SSIM) [60], Learned Perceptual Image Patch Similarity (LPIPS) [51], Root Mean Square Error (RMSE), and Correlation Coefficient (CC).

PSNR reflects pixel-level reconstruction accuracy, UIQI measures overall image similarity, and SSIM focuses on local structural similarity. LPIPS quantifies perceptual similarity based on deep feature representations. RMSE and CC evaluate global statistical consistency. The combination of these complementary metrics provides a more comprehensive and robust evaluation of the reconstruction quality from multiple perspectives.

## III. Experiments and Results

*A. Dataset*

A dataset, referred to as the SLC-off Antarctic dataset (SLCANT), is constructed to train and evaluate the proposed DiffGF framework. It consists of 3,139 image patches of 256×256 pixels, collected from 10 representative regions across Antarctica (Fig. A1), with an emphasis on structurally complex and diverse ice shelf areas, while also including ice sheet regions.

These samples cover a wide range of Antarctic surface types and geomorphological features, including snow- and ice-covered surfaces, coastlines, bare rock, blue ice, melt ponds, ice streams, and fractures (e.g., crevasse fields and rifts), ensuring high structural diversity.

Constrained by polar nights, all imagery was acquired during the austral summer. This does not limit the applicability of the dataset, as virtually all usable optical observations in Antarctica are obtained during this period, which is consistent with the inherent characteristics of Antarctic optical remote sensing.

Additional details of the selected images are provided in Table A1. For Shackleton Ice Shelf, two spatially non-overlapping acquisitions were used because a single acquisition could not provide sufficient cloud-free coverage over the selected region. The SLCANT dataset is divided into training and test sets, with 80% (2,514 images) for training and 20% (625 images) for testing. Image patches are randomly split within each region to improve the diversity and representativeness of Antarctic surface characteristics in both the training and test sets. All patches are spatially non-overlapping.

To generate simulated SLC-off images with corresponding ground truth, cloud-free Landsat 8 OLI top-of-atmosphere (TOA) products were selected. Missing-data masks were derived from co-located Landsat 7 ETM+ TOA products acquired under SLC-off conditions and were subsequently applied to the Landsat 8 images (Fig. 3). This process generates realistic SLC-off patterns while preserving the original complete Landsat 8 data as reference ground truth. Accordingly, all training and evaluation are performed on simulated SLC-off images derived from Landsat 8, with the original Landsat 8 images serving as ground truth. Landsat 7 imagery is used solely to provide realistic SLC-off masks in the SLCANT dataset. The images in Fig. 3 also illustrate pronounced surface changes over the image acquisition period, which are substantially stronger than those typically

TABLE I

QUANTITATIVE METRICS ON THE SLCANT TEST SET

| Type | Band | Metric | | | | | |
|---|---|---|---|---|---|---|---|
| | | PSNR↑ | UIQI↑ | SSIM↑ | CC↑ | RMSE↓ | LPIPS↓ |
| SLC-off (full image) | R | 10.6830 | 0.1660 | 0.6415 | 0.2130 | 0.3151 | - |
| | G | 10.7662 | 0.1569 | 0.6400 | 0.2018 | 0.3114 | - |
| | B | 9.9085 | 0.1479 | 0.6376 | 0.1884 | 0.3433 | - |
| | Mean | 10.4525 | 0.1569 | 0.6397 | 0.2011 | 0.3233 | 0.5227 |
| DiffGF restoration (full image) | R | 50.5647 | 0.9738 | 0.9894 | 0.9743 | 0.0060 | - |
| | G | 51.4775 | 0.9700 | 0.9905 | 0.9706 | 0.0053 | - |
| | B | 51.9015 | 0.9680 | 0.9913 | 0.9687 | 0.0048 | - |
| | Mean | 51.3146 | 0.9706 | 0.9904 | 0.9712 | 0.0054 | 0.0176 |
| DiffGF restoration (mask only) | R | 41.8837 | 0.7731 | 0.9562 | 0.7798 | 0.0160 | - |
| | G | 42.8022 | 0.7522 | 0.9609 | 0.7582 | 0.0143 | - |
| | B | 43.2120 | 0.7421 | 0.9633 | 0.7490 | 0.0129 | - |
| | Mean | 42.6327 | 0.7558 | 0.9601 | 0.7623 | 0.0144 | - |

observed in most non-polar environments and are representative of the persistent and highly dynamic nature of Antarctic surfaces.

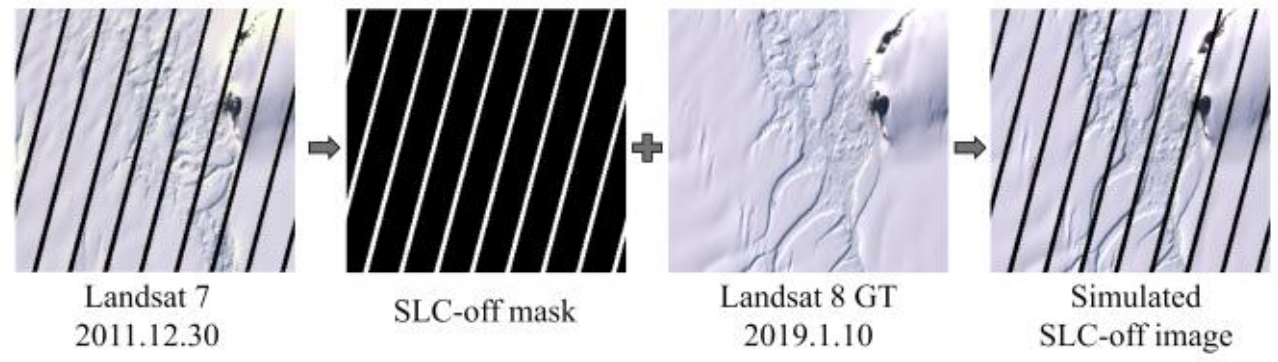

**Fig. 3.** Illustration of constructing simulated SLC-off images in the SLCANT dataset. The example image is centered at 110°27′55″ W, 75°11′12″ S (Crosson Ice Shelf, West Antarctica). An SLC-off mask derived from a Landsat 7 ETM+ image acquired on December 30, 2011, was applied to a cloud-free, co-located Landsat 8 OLI image acquired on January 10, 2019, to generate a simulated SLC-off image with known ground truth.

All data were obtained via the Google Earth Engine (GEE) platform. Cloud and cloud-shadow pixels were masked using the cloud and cloud-shadow flags in the 'QA_PIXEL' quality assessment band provided with the Landsat Collection 2 products in GEE, and only cloud-free patches were retained. The red, green, and blue (RGB) bands (Bands 4, 3, and 2 for Landsat 8; Bands 3, 2, and 1 for Landsat 7) were selected and converted to 8-bit unsigned integers (uint8) in GEE. Due to the lack of the Tier 1 products over Antarctica, Collection 2 Tier 2 products were used for both Landsat 8 and Landsat 7. All images were exported in the WGS84 Antarctic Polar Stereographic projection (EPSG: 3031).

### *B. Implementation Details*

The proposed DiffGF framework is implemented using PyTorch. All training is conducted on an NVIDIA A100 GPU within a shared high-performance computing (HPC) environment, where GPU resources are not exclusively allocated.

The diffusion model is trained on the SLCANT dataset with a batch size of 8 for 200 epochs. Four sampling steps are adopted in the diffusion process. The loss function in Eq. (3) uses $\lambda$=10. AdamW is adopted as the optimizer, with an initial learning rate of $10^{-4}$. A two-stage learning rate schedule is applied: constant during the first 50 epochs, followed by cosine decay to $5 \times 10^{-5}$. Automatic Mixed Precision (AMP) is optionally enabled to improve training efficiency. During training, random horizontal and vertical flips are applied for data augmentation.

As described above, MGHNet (Fig. 1c) is trained using the processed ground-truth images from the SLCANT dataset, which are encoded and decoded by the pretrained VQGAN, along with their corresponding masks. Training is conducted with a batch size of 32 for 1000 epochs. The loss function in Eq. (5) uses $\lambda$=200 and $A_{min}$=10, which were empirically selected to enhance harmonization within the masked regions. The encoder follows the Swin Transformer v2 base configuration with a patch size of 4 and a window size of 8, selected to match the 256 × 256 input patch size. The initial and final convolutional layers use 3 × 3 kernels with padding 1. Following the encoder, the channel dimensions are progressively reduced from 256 to 128, 64, and 32. Three PixelShuffle blocks with upscale factors of 4, 4, and 2 are configured to progressively recover the spatial resolution from the encoder output. A staged learning rate schedule is applied throughout the 1000 epochs: it linearly increases from $10^{-5}$ to $10^{-4}$ for the first 20 epochs, remains constant at $10^{-4}$ from epoch 20 to 100, then decays to $5 \times 10^{-5}$ between epochs 100 and 500, and further reduces to $2 \times 10^{-5}$ from epoch 500 to 1000. During training, random rotations are applied as data augmentation.

### *C. Results on SLCANT Dataset*

#### *1) Quantitative Results*

Table I presents the quantitative performance of the proposed DiffGF framework on the test set of the SLCANT dataset. To comprehensively evaluate reconstruction quality, metrics are computed both within the missing (masked) regions and over the entire image. For each band, PSNR, UIQI, SSIM, CC, and RMSE are calculated, and the average

TABLE II

PERFORMANCE OF DIFFERENT DIFFGF VARIANTS ON THE SLCANT TEST SET. ALL METRICS EXCEPT LPIPS ARE COMPUTED AS THE MEAN OF PER-BAND VALUES WITHIN THE MASKED REGIONS.

| Method | Metric | | | | | |
|---|---|---|---|---|---|---|
| | **PSNR↑** | **UIQI↑** | **SSIM↑** | **CC↑** | **RMSE↓** | **LPIPS/full↓** |
| DiffGF | **42.6327** | **0.7558** | 0.9601 | **0.7623** | **0.0144** | **0.0176** |
| Diffusion-only | 41.1953 | 0.7322 | 0.9577 | 0.7407 | 0.0154 | 0.0421 |
| Sampling step: 2 | 42.5697 | 0.7539 | 0.9591 | 0.7602 | 0.0146 | 0.0180 |
| Sampling step: 8 | 42.6143 | 0.7556 | **0.9604** | 0.7623 | 0.0144 | 0.0176 |
| Input: Replace Δ with y0 | 42.1895 | 0.7505 | 0.9597 | 0.7571 | 0.0147 | 0.0181 |
| ResBlock-Variant 1 | 42.4559 | 0.7507 | 0.9597 | 0.7575 | 0.0145 | 0.0177 |
| ResBlock-Variant 2 | 42.1500 | 0.7430 | 0.9592 | 0.7502 | 0.0147 | 0.0179 |

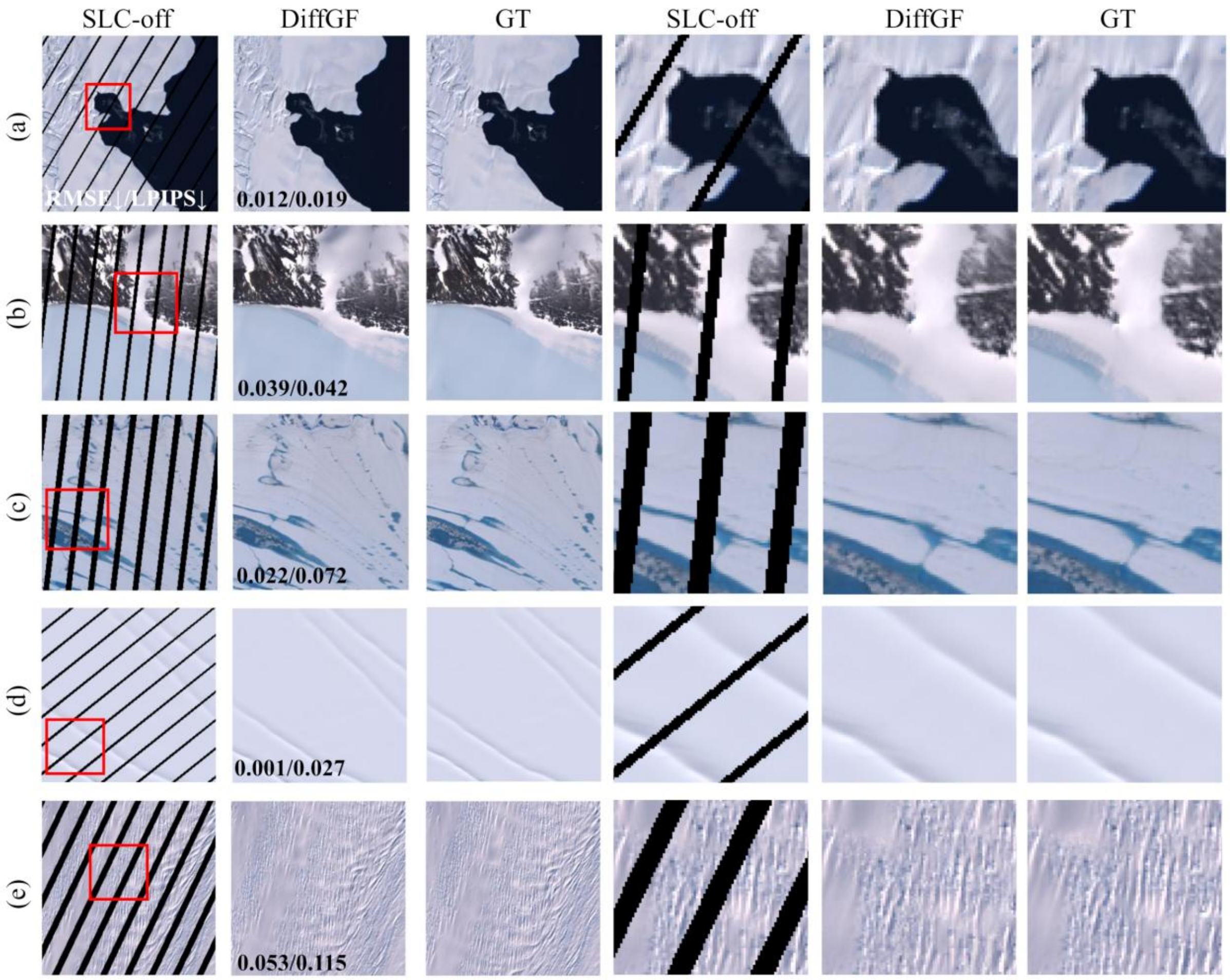

**Fig. 4.** Restoration results of DiffGF on test images from the SLCANT dataset. From left to right: SLC-off input images, reconstructed results by DiffGF, and ground truth images (GT), and zoomed-in views of the red-boxed regions. (a) Acquired on February 16, 2019, over Shackleton Ice Shelf. (b)-(c) Acquired on January 20, 2020, over Amery Ice Shelf. (d) Acquired on December 29, 2020, over Larsen C Ice Shelf. (e) Acquired on February 28, 2018, over Pine Island Ice Shelf. The RMSE and LPIPS values computed over the entire image are annotated in each subfigure.

values across all bands are also reported as the overall performance.

As shown in the table, the reconstructed masked regions exhibit strong consistency with the ground truth, indicating the effectiveness of DiffGF in recovering missing information. Correspondingly, the overall image quality is significantly improved compared to the original SLC-off inputs, further demonstrating the capability of DiffGF in high-fidelity reconstruction.

As a non-reference restoration method, quantitative image quality metrics provide an objective measure of overall reconstruction fidelity. However, for SLC-off restoration in Antarctica, image quality metrics alone are insufficient to fully characterize the effectiveness of a non-reference method designed for remote sensing applications. The goal of this study is not only to improve numerical similarity, but also to preserve structural continuity, semantic consistency, and physical plausibility, which are critical for visual interpretation and downstream analysis. Therefore, qualitative evaluations and downstream application experiments are further presented in the following sections to provide a more comprehensive assessment of DiffGF.

*2) Qualitative Results*

Representative visual examples from the SLCANT test set processed by the proposed DiffGF framework are presented in Fig. 4. The reconstructed images across diverse surface types and geomorphological features exhibit high visual quality and strong consistency with the ground truth. Missing regions are effectively filled with realistic textures and structures, with smooth and natural transitions observed at the boundaries between reconstructed and known regions.

Fine-scale details, such as edges and linear features, are effectively recovered. For example, reconstructed coastlines (Fig. 4(a)), melt pond boundaries (Fig. 4(c)), and large-scale crevasses (Fig. 4(d)) show high structural fidelity to the ground truth. In addition to bright ice and snow surfaces, darker bare rock regions (Fig. 4(b)) are well reconstructed, with clear boundaries that are highly consistent with the ground truth. Moreover, complex and chaotic crevasse fields (Fig. 4(e)) are successfully recovered, indicating the capability of DiffGF to handle highly heterogeneous and structurally complex Antarctic surfaces, which is critical for downstream remote sensing interpretation and analysis.

Despite the presence of parallel stripe-shaped missing patterns, the reconstructed textures closely resemble the ground truth, preserve fine spatial structures, and maintain high structural fidelity. These results demonstrate the robustness and generalization capability of DiffGF across diverse Antarctic regions and surface conditions.

### *D. Ablation Study*

To assess the effectiveness of the proposed DiffGF framework, ablation studies are performed on the SLCANT dataset under identical training configurations to ensure fair comparisons. All metrics in Table II, except for LPIPS, are calculated separately for each band within the reconstructed (masked) regions and then averaged across bands to evaluate gap-filling performance, whereas LPIPS is calculated over the entire image to assess overall perceptual similarity.

*1) Contribution of the Mask-Guided Harmonization Network (MGHNet)*

To assess the contribution of the refinement stage in the proposed two-stage DiffGF framework, we conduct an ablation study by removing MGHNet and relying solely on the diffusion model for reconstruction. As shown in Table II, the inclusion of MGHNet consistently improves reconstruction quality across all metrics. This demonstrates that the latent-space diffusion model alone is insufficient to achieve high-quality reconstruction, challenged by the spatial discontinuities caused by the stripe-patterned missing regions in Landsat 7 SLC-off imagery.

Fig. 5 further illustrates this effect. Without MGHNet, the reconstructed images exhibit visible artifacts and abrupt transitions at the boundaries between reconstructed and known regions. The SLC-off stripe artifacts are highlighted by arrows in Fig. 5(b) and (d), as the overall high surface brightness and low contrast make visual discrimination more difficult. In contrast, the full DiffGF pipeline with MGHNet produces smoother and more coherent transitions, substantially enhancing local continuity and spatial consistency.

Both quantitative and qualitative results demonstrate the effectiveness of the MGHNet and underscore the necessity of the proposed two-stage design. While latent-space diffusion enables computational efficiency, the subsequent refinement step with MGHNet significantly improves spatial and spectral consistency.

This modular design offers a practical trade-off between reconstruction performance and computational cost, making it particularly suitable for resource-constrained or high throughput applications. It may also serve as a reference for future designs of generative frameworks in remote sensing tasks under limited-resource scenarios.

*2) Impact of Sampling Steps in the Diffusion Process*

To assess the impact of the number of sampling steps in the diffusion process on the final restoration results, we conduct an ablation study comparing restoration performance with 2, 4, and 8 sampling steps. As shown in Table II, the 4-step setting (DiffGF) achieves the best overall performance, with only a slightly lower SSIM value than the 8-step setting. This result indicates that, for the SLC-off restoration task considered in this study, 4 sampling steps provide a good balance between restoration fidelity and structural preservation. Therefore, the number of sampling steps in the diffusion process of DiffGF is set to 4. Notably, further increasing the number of sampling steps does not lead to an evident performance gain, suggesting that the marginal benefit of additional sampling steps is limited under the current task and model settings.

*3) Effect of Input Guidance in MGHNet*

To analyze the impact of input guidance on the performance of MGHNet, we investigate different input configurations by comparing two designs: one that incorporates pixel-wise difference within unmasked regions ($\Delta$) between $\hat{x}_0$ and $y_0$, and another that directly uses $y_0$, both alongside $\hat{x}_0$ and the mask. The quantitative results are shown in Table II. As the metrics indicate, explicitly incorporating the pixel-wise difference leads to better performance than directly using $y_0$. This demonstrates that the difference map serves as a more informative and task-relevant input, effectively guiding the network to learn the refinement task. This design is inherently consistent with MGHNet's objective of predicting residual corrections.

*4) Impact of the Block Design in the Decoder of MGHNet*

To evaluate the impact of the customized residual block (ResBlock) used in MGHNet, we conduct ablation experiments comparing different decoder block designs. In addition to the designed ResBlock (Fig. 6(a)), two alternative variants are examined. The first variant (Fig. 6(b)) incorporates Batch Normalization (BN) layers into ResBlock to assess the influence of normalization. The second variant adopts a classic ResNet-style block[56] (Fig. 6(c)) that includes BN and employs ReLU as the activation functions in both intermediate and output stages.

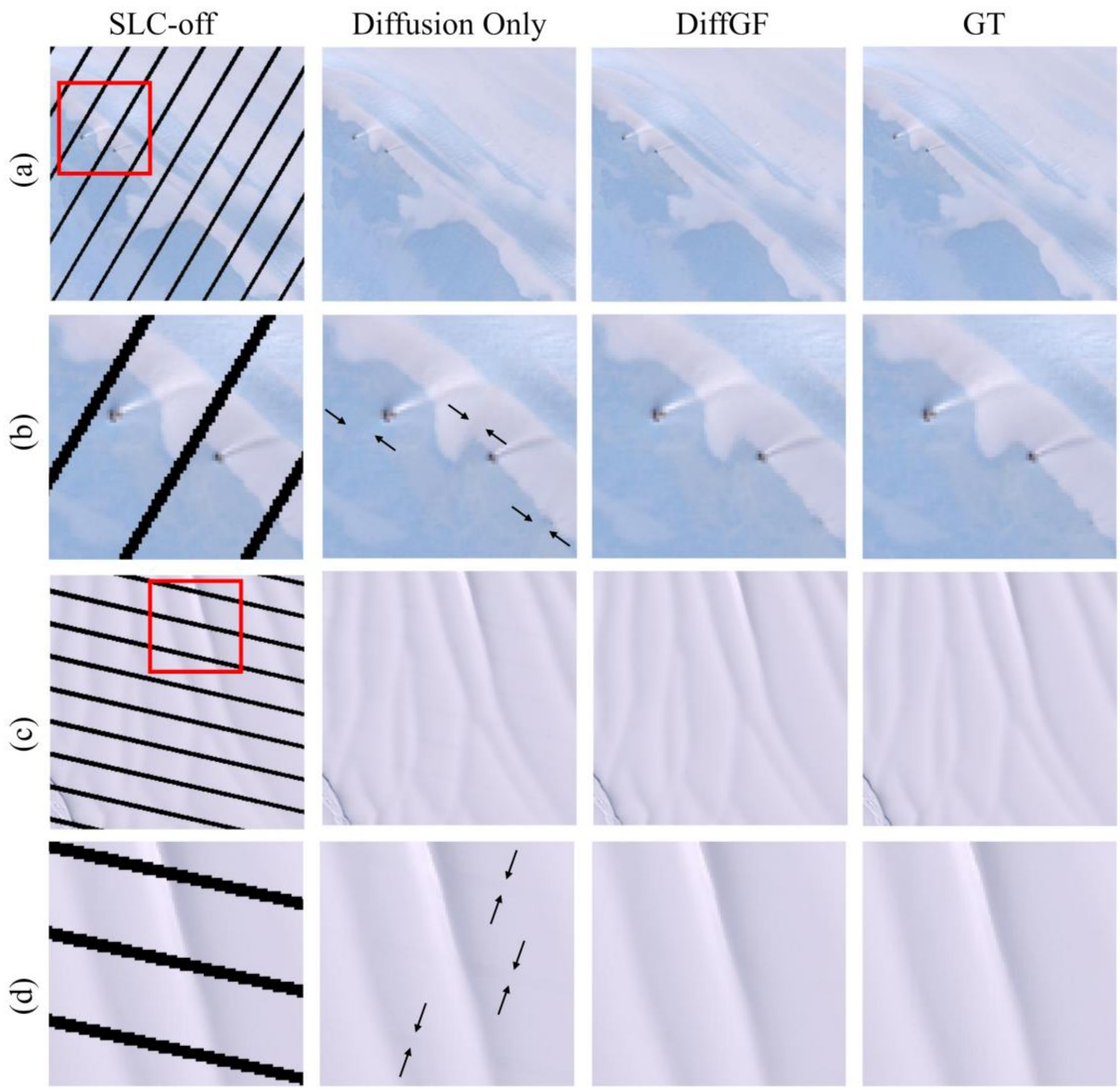

**Fig. 5.** Comparison of DiffGF with and without MGHNet on test images from the SLCANT dataset. From left to right: SLC-off input images, reconstructed results using the diffusion model only, DiffGF results, and ground truth images. (a) Acquired on February 16, 2019, over Shackleton Ice Shelf. (b) Zoomed-in views of the red-boxed region in (a). (c) Acquired on January 7, 2021, over Brunt Ice Shelf. (d) Zoomed-in views of the red-boxed region in (c).

Quantitative results on the SLCANT test set are summarized in Table II. The MGHNet equipped with ResBlock consistently outperforms the other two designs across all metrics. These results highlight the superior suitability of designed ResBlock for the pixel-level correction task. This improvement can be attributed to two key design choices: (1) removing BN may help preserve local features and avoid normalization artifacts; and (2) using tanh instead of ReLU allows for finer control over output range, which is beneficial for restoring subtle image details.

These findings support the effectiveness of the ResBlock design and demonstrate its contribution to enhancing the overall reconstruction quality within the DiffGF framework.

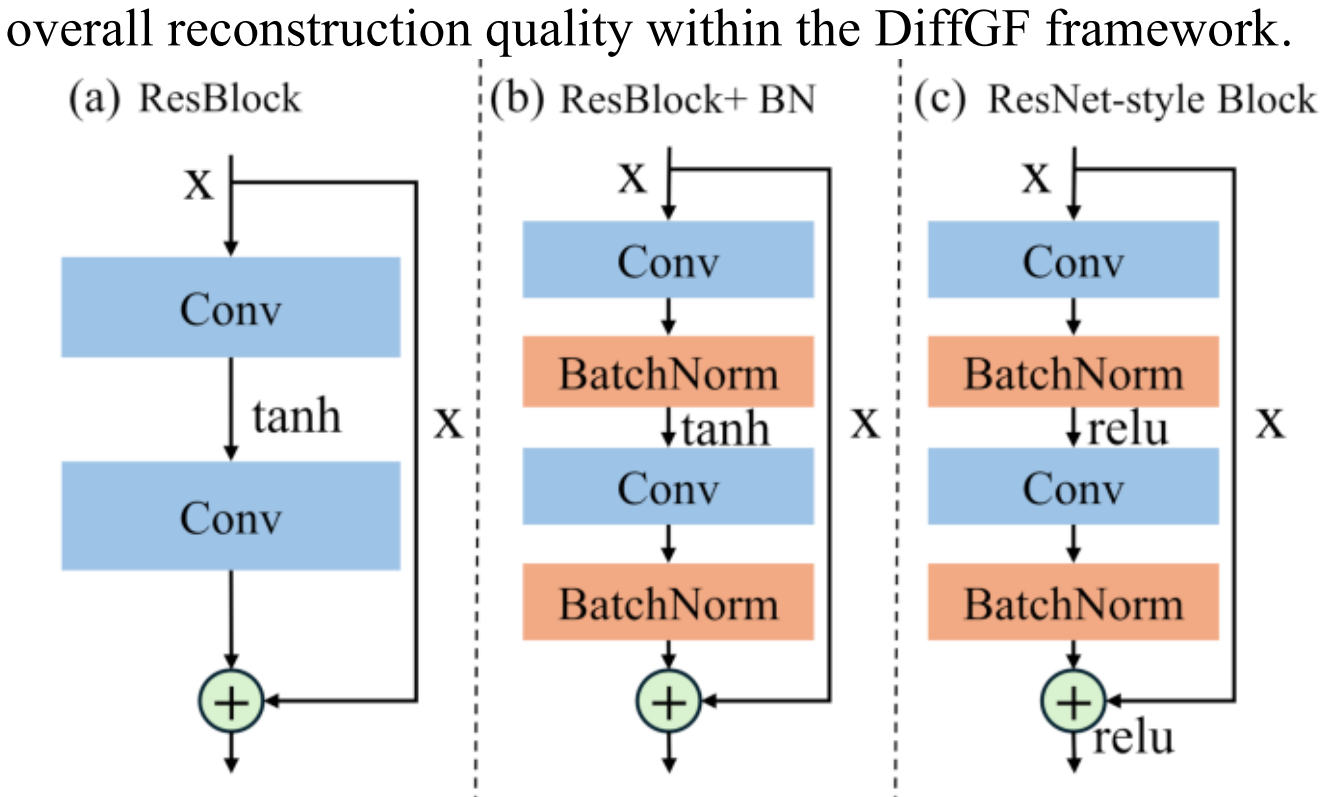

**Fig. 6.** Architectures of the residual block used in the ablation study for the MGHNet decoder. (a) The designed ResBlock. (b) Variant 1: ResBlock with Batch Normalization (BN). (c) Variant 2: Classical ResNet- style block following [56].

TABLE III
QUANTITATIVE EVALUATION OF CREVASSE-CLASS SEGMENTATION RESULTS BASED ON RESTORED IMAGERY (EVALUATED AGAINST THE SEGMENTATION RESULTS FROM THE ORIGINAL COMPLETE IMAGERY).

| Method/Data | IoU↑ | OA↑ | Recall↑ | Precision↑ | F1↑ |
|---|---|---|---|---|---|
| SLC-off | 0.4688 | 0.9205 | 0.6157 | 0.6626 | 0.6383 |
| Gapfill | 0.8261 | 0.9794 | 0.8592 | 0.9554 | 0.9048 |
| LaMa [22] | 0.8941 | 0.9874 | **0.9357** | 0.9527 | 0.9441 |
| RePaint [32] | 0.9004 | 0.9882 | 0.9354 | 0.9601 | 0.9476 |
| DiffGF | **0.9019** | **0.9885** | 0.9286 | **0.9691** | **0.9484** |

TABLE IV
QUANTITATIVE IMAGE QUALITY METRICS COMPUTED OVER THE FULL RESTORED TEST IMAGE USING DIFFERENT METHODS.

| Method | PSNR↑ | UIQI↑ | SSIM↑ | CC↑ | RMSE↓ | LPIPS↓ |
|---|---|---|---|---|---|---|
| Gapfill | 34.6037 | 0.9928 | 0.9646 | 0.9928 | 0.0188 | 0.1009 |
| LaMa [22] | 35.4617 | 0.9942 | 0.9623 | 0.9942 | 0.0170 | 0.1397 |
| RePaint [32] | 34.5383 | 0.9927 | 0.9648 | 0.9927 | 0.0190 | 0.0659 |
| DiffGF | 35.0703 | 0.9935 | 0.9659 | 0.9936 | 0.0178 | 0.0602 |

### *E. Downstream Application: Crevasse Segmentation*

#### *1) Crevasse Segmentation Method*

To evaluate the semantic fidelity of reconstructed images and their suitability for practical applications, crevasse segmentation is selected as a representative downstream task for Antarctic imagery.

Specifically, a cloud-free Landsat 8 OLI image of Larsen B Ice Shelf, acquired on December 21, 2021 is selected as an independent test scene. Larsen B Ice Shelf contains diverse crevasse types, including linear crevasses and crevasse fields, providing a representative setting for assessing the generalization capability and application-oriented reliability. To simulate SLC-off conditions with known ground truth, an SLC-off mask extracted from a co-located Landsat 7 ETM+ image acquired on February 20, 2012, is applied to the Landsat 8 image, following the same procedure used in constructing the SLCANT dataset. The simulated SLC-off image covers an area of 53.8 km × 76.8 km (Fig. 7(a)) and is divided into 70 non-overlapping patches of 256 × 256 pixels for evaluation.

In addition, a crevasse segmentation dataset comprising 276 image patches of 512×512 pixels is constructed from cloud-free Landsat 8 imagery collected across seven different ice shelves during the austral summer of 2020. Manual delineation is assisted with multi-source remote sensing data, including Sentinel-1 SAR and ICESat-2 altimetry. These patches are used to train and evaluate a semantic segmentation model. The trained model is then applied to the reconstructed images generated by DiffGF to assess its ability in recovering and preserving semantic information.

The SegFormer-B2 network [61], a transformer-based semantic segmentation network, is adopted as the classifier. The 276 image patches are split into a training set (224) and a test set (52), with no spatial overlap with the Larsen B Ice Shelf test image, ensuring independent evaluation. The model is trained for 200 epochs using AdamW optimizer (learning rate = $1\times10^{-4}$) with a batch size of 8. A combined loss of Binary Cross-Entropy and Dice Loss, is employed to address class imbalance. The trained model achieves a crevasse-class recall of 0.8406 and precision of 0.8165, and a mean Intersection over Union (IoU) of 0.8451 on the test set, demonstrating its effectiveness for Antarctic crevasse segmentation.

#### *2) Crevasse Segmentation Results*

To evaluate the restoration capability of DiffGF, crevasse segmentation results based on DiffGF-restored imagery are compared with those obtained from the original complete Landsat 8 image. The restored SLC-off image is displayed in Fig. 7(a), where DiffGF effectively recovers the missing stripes with high quality. As illustrated in Figs. 8(a) and 9(a), structural patterns and textures of crevasses are faithfully restored by DiffGF.

Quantitative segmentation results are reported in Table III. Crevasse segmentation based on DiffGF-restored imagery achieves an overall accuracy (OA) of 0.9885, while yielding a crevasse-class IoU of 0.9019, a recall of 0.9286, a precision of 0.9691, and an F1-score of 0.9484, when evaluated against the segmentation results derived from the original complete image. As shown in Figs. 8 and 9, the effective recovery of semantic structures enables crevasse segmentation results that closely resemble those obtained from the original complete image. In contrast, segmentation results derived directly from the SLC-off imagery (Fig. 7(b)) fail to provide visually meaningful crevasse segmentation.

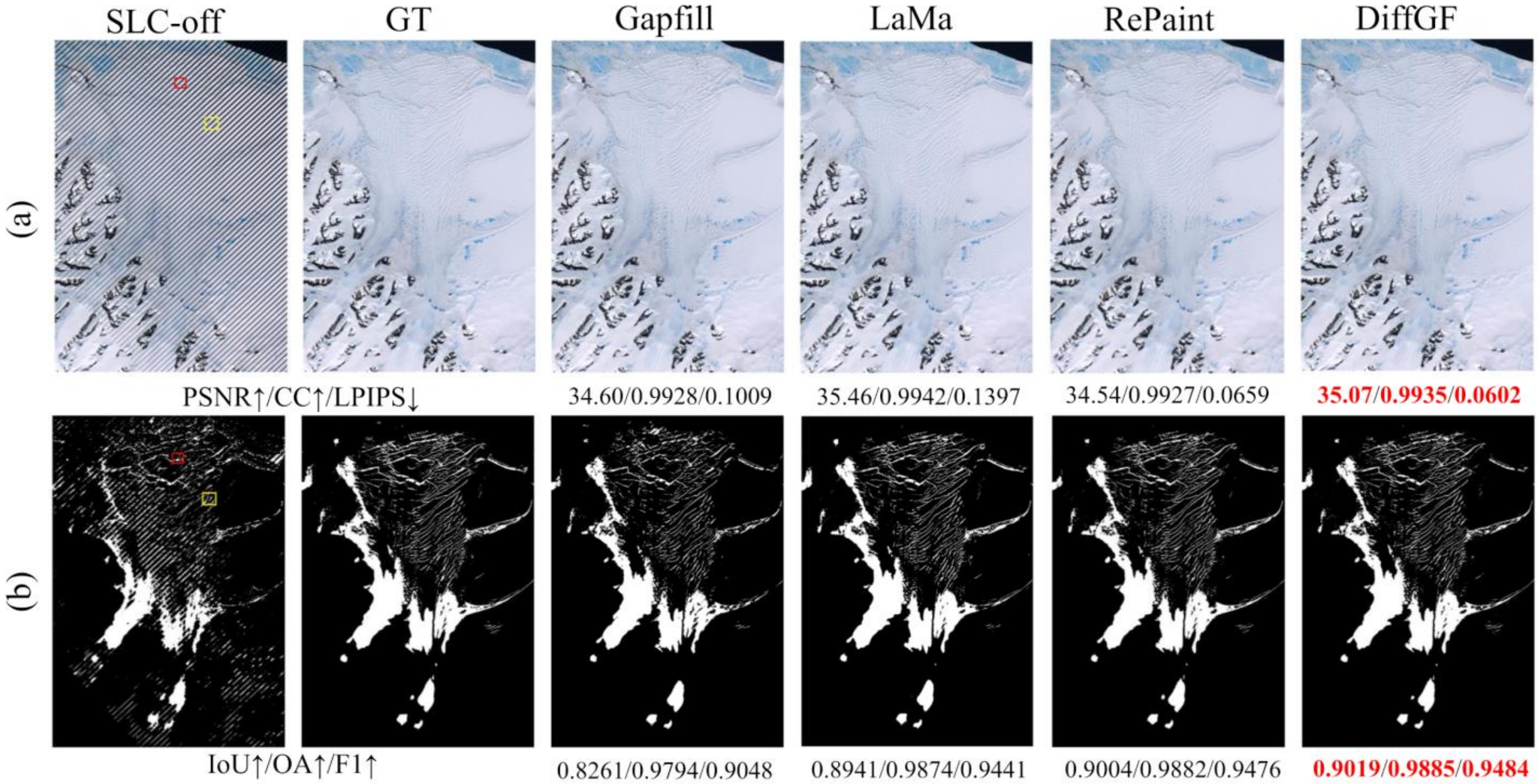

**Fig. 7.** Reconstruction and downstream crevasse segmentation results on the simulated SLC-off Larsen B Ice Shelf test image. (a) Reconstructed images generated by different methods. (b) Corresponding crevasse segmentation results.

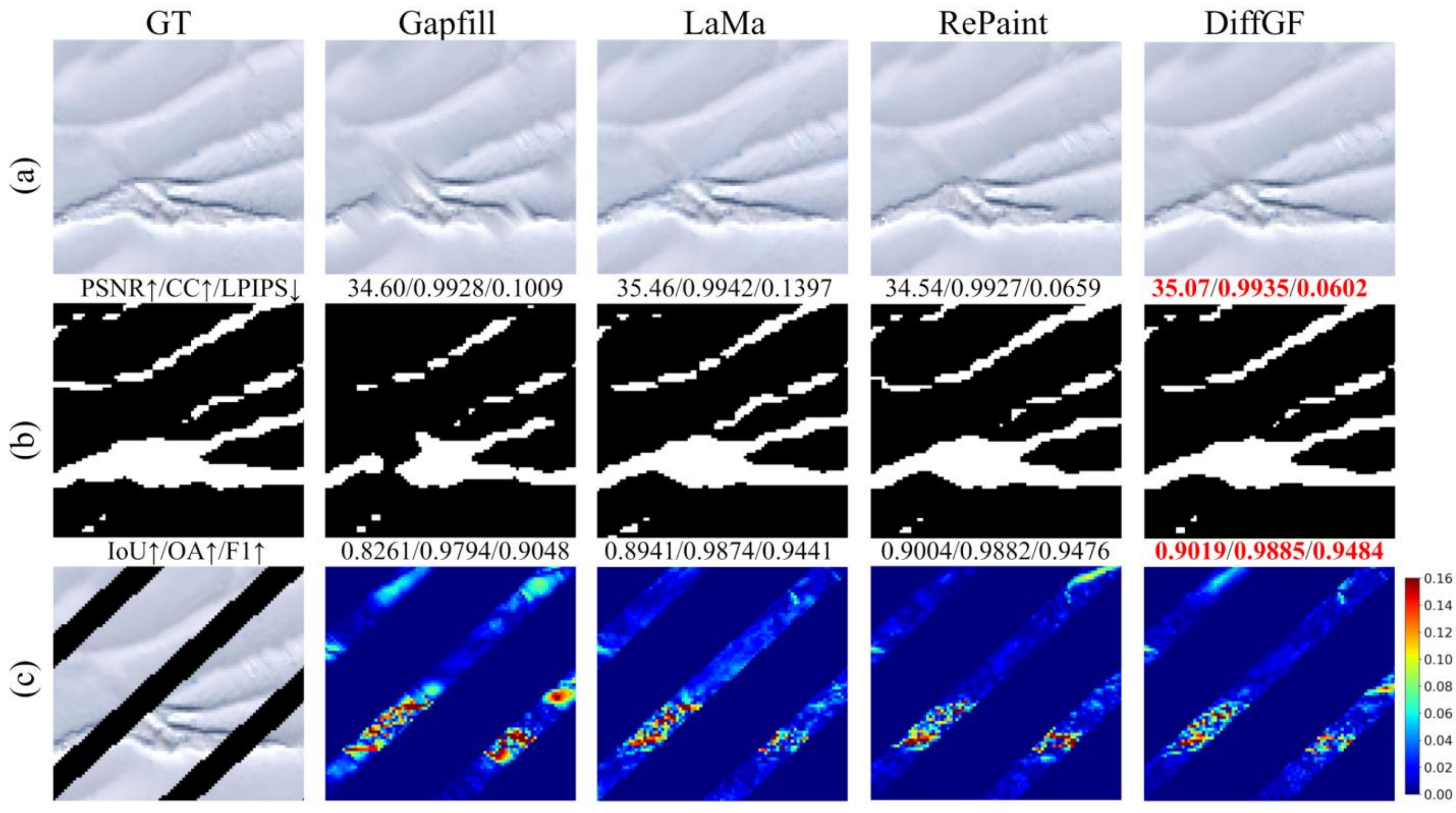

**Fig. 8.** Zoomed-in views of the red-boxed regions in Fig. 7, showing detailed reconstruction results (a), crevasse segmentation results (b), and the SLC-off input together with corresponding error maps of the reconstructed images (c).

These results demonstrate that the crevasse segmentation based on DiffGF-restored imagery is remarkably close to that based on the complete ground truth image, indicating that DiffGF effectively recovers and preserves semantic information critical for crevasse mapping.

This application-oriented evaluation provides a task-driven perspective on reconstruction performance, complementing the aforementioned image quality metrics. Together, these results further evaluate the performance of DiffGF in semantic preservation and provide supportive evidence for its practical utility in addressing the Landsat 7 ETM+ SLC-off restoration problem in Antarctic remote sensing applications such as crevasse mapping.

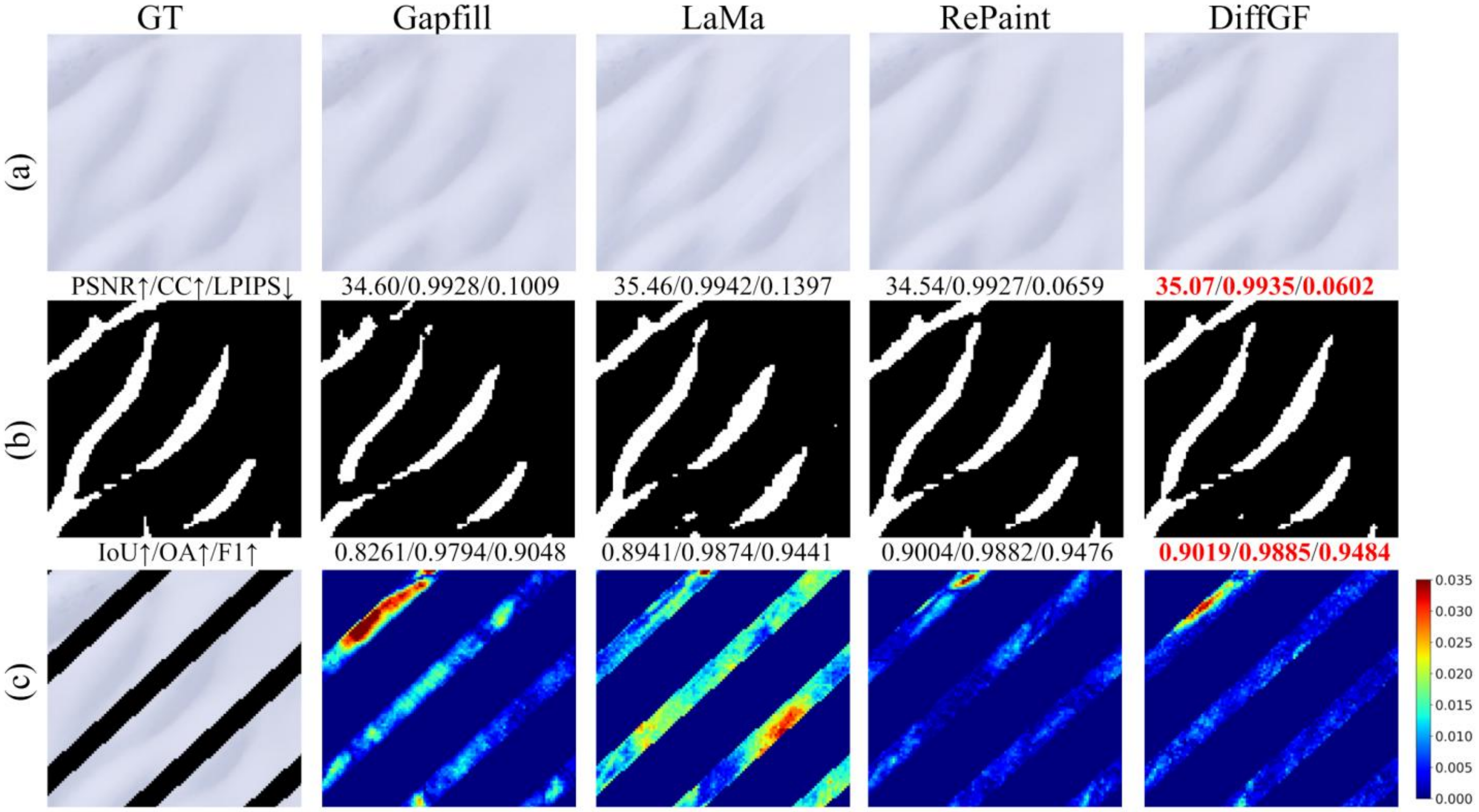

**Fig. 9.** Zoomed-in views of the yellow-boxed regions in Fig. 7, showing detailed reconstruction results (a), crevasse segmentation results (b), and the SLC-off input together with corresponding error maps of the reconstructed images (c).

*F. Comparison with Existing Methods*

To evaluate the proposed DiffGF framework against existing methods, comparative experiments are conducted on the simulated Larsen B Ice Shelf SLC-off image described in the previous section.

*1) Comparative Methods*

We compare the DiffGF framework with three representative non-reference methods spanning both SLC-off restoration and general image inpainting:

1. Gapfill: the ENVI Landsat Gapfill tool that estimates missing pixels by interpolating values from surrounding known regions.

2. LaMa [22]: a GAN-based method that uses Fast Fourier convolutions (FFCs) to achieve an image-wide receptive field to produce high-fidelity reconstruction.

3. RePaint [32]: a diffusion-based method that leverages a pretrained unconditional DDPM as the generative prior and performs mask-conditioned reconstruction.

To ensure a fair comparison, both LaMa and RePaint were fine-tuned on the SLCANT dataset using their official implementations, initialized from their respective pretrained checkpoints. LaMa was fine-tuned for 100 epochs with a batch size of 8, and the DDPM used in RePaint was fine-tuned for 50k iterations with a batch size of 2.

*2) Quantitative and Qualitative Comparison*

Table IV and Fig. 7 summarize the quantitative and qualitative restoration results of DiffGF and the comparative methods on the simulated SLC-off Larsen B Ice Shelf test image.

As reported in Table IV, all the metrics are computed over the full image, with the average values across all bands reported as the overall performance, except for LPIPS. DiffGF achieves consistently strong performance across all criteria. Notably, DiffGF attains the best LPIPS score among all comparative methods, indicating superior perceptual similarity and high-level structural fidelity. While LaMa achieves marginally higher scores in distortion-based metrics, the differences between DiffGF and LaMa remain small, suggesting comparable pixel-level reconstruction quality.

A notable phenomenon is observed for Gapfill. Although the Gapfill-restored image exhibits the poorest visual quality, characterized by visually blurred and structurally oversmoothed reconstructions (Figs. 8(b) and 9(b)), its distortion-based metric scores remain comparable to those of the other comparative methods. This inconsistency between distortion-based metrics and visual reconstruction quality suggests that conventional distortion-based image quality metrics alone may be insufficient to comprehensively evaluate non-reference SLC-off restoration performance.

Qualitative comparisons in Figs. 8 and 9 further demonstrate that DiffGF produces more semantically coherent reconstructions while preserving structural continuity with the known regions. In contrast, the LaMa-restored result exhibits blurred crevasse edges in Fig. 8, and visible discontinuities and boundary artifacts between reconstructed and known regions in Fig. 9. The RePaint-restored result also shows deviations of crevasse boundaries from the ground truth in Fig. 8. Overall, crevasse structures, textures, and boundaries are more faithfully recovered by DiffGF, showing closer agreement with the ground-truth image than the comparative methods.

To further visualize local reconstruction fidelity, pixel-wise absolute differences between the reconstructions and the ground-truth image are computed and normalized to [0,1], as shown in Figs. 8(c) and 9(c), illustrating that DiffGF-restored imagery exhibits smaller differences from the original ground-truth imagery. Taken together, these results demonstrate that DiffGF outperforms the comparative methods in both image fidelity and semantic recovery.

*3) Downstream Application Evaluation*

To further assess the practical utility of different restoration methods, downstream crevasse segmentation results are reported in Table III and illustrated in Figs. 8(b) and 9(b). DiffGF achieves the best overall performance, with the highest IoU, OA, precision, and F1-score, and only a marginal difference in recall compared to LaMa and RePaint. The higher IoU and precision indicate that DiffGF produces more accurate semantic structures for crevasse delineation (Figs. 8(b) and 9(b)), while more effectively suppressing restoration-induced false positives. The poor segmentation performance based on Gapfill-restored imagery is consistent with previous discussions, further suggesting that image quality metrics alone are insufficient to comprehensively evaluate non-reference restoration methods.

Although LaMa achieves slightly higher scores on several distortion-based image metrics, segmentation performance based on DiffGF-restored imagery is better in this downstream task, suggesting that DiffGF can better preserve task-relevant semantic information for crevasse mapping. This further supports the value of application-oriented evaluation for a comprehensive and fair assessment of non-reference SLC-off restoration methods.

*4) Inference Efficiency of Diffusion-Based Methods*

Since DiffGF and RePaint are both diffusion-based approaches, we provide an illustrative comparison of inference efficiency between these two methods. The experiments are conducted on the same device (NVIDIA GeForce RTX 4060 Laptop GPU). Reconstructing the same $256 \times 256$ SLC-off image takes approximately 423 seconds with RePaint, whereas DiffGF requires only 0.4 seconds. This difference of approximately three orders of magnitude stems from both the latent-space diffusion in DiffGF and the lightweight sampling strategy (4 steps versus 250 steps in RePaint), indicating the efficiency advantage of our framework. We emphasize that this is not a rigorous benchmarking experiment, as inference efficiency is not the main focus of this study. Nevertheless, this comparison highlights the practical computational advantage of DiffGF.

Overall, DiffGF demonstrates competitive and well-balanced performance among non-reference methods. Both image quality and application-oriented evaluations indicate that DiffGF produces visually plausible reconstructions, providing a useful solution for supporting Antarctic remote sensing applications such as crevasse mapping.

## IV. Discussion

*A. Extension to Other Scenarios*

This study is primarily motivated by the challenge of Landsat 7 SLC-off imagery restoration in Antarctica. In this region, historical optical observations are inherently scarce, surfaces undergo rapid changes, suitable reference data are often unavailable, and existing imagery archives are exceptionally valuable for long-term analysis.

However, the applicability of the proposed DiffGF framework may not be limited to the Antarctic environment and could be extended to other regions. Nevertheless, it is worth noting that in many non-polar regions, usable reference imagery is often available due to more stable surface conditions and the absence of polar nights, and reference-based restoration methods may therefore remain more suitable.

Despite this, DiffGF remains valuable in other specific situations where reference data are unavailable, such as historical post-disaster mapping and other rapid-change scenarios. Furthermore, the proposed non-reference diffusion-based framework can potentially be extended to other remote sensing data reconstruction tasks, such as cloud removal.

*B. Dataset*

The SLCANT dataset is constructed using only RGB bands, focusing on natural color imagery. RGB imagery provides an important data source for Antarctic surface monitoring tasks that rely strongly on visible structural and textural information, such as crevasse mapping, melt pond detection, and calving-front delineation. For these tasks, restoring SLC-off gaps can improve the spatial continuity and usability of historical Landsat 7 observations, particularly for time-series monitoring.

While this design supports visual interpretability and facilitates the evaluation of structural and semantic reconstruction capability, it also constrains the available spectral diversity. We note that this study focuses on the RGB domain, and its extension to multispectral restoration may require further evaluation using additional spectral bands to support applications that rely on multispectral information. Future datasets can extend the band selection to accommodate application scenarios requiring information from additional wavelengths, such as near-infrared, shortwave infrared, thermal, and panchromatic bands. Incorporating multispectral inputs may further enhance the capability of non-reference restoration methods to capture richer spectral information and support a wider range of remote sensing tasks.

It should be noted that the SLCANT dataset was constructed using complete Landsat 8 images to simulate

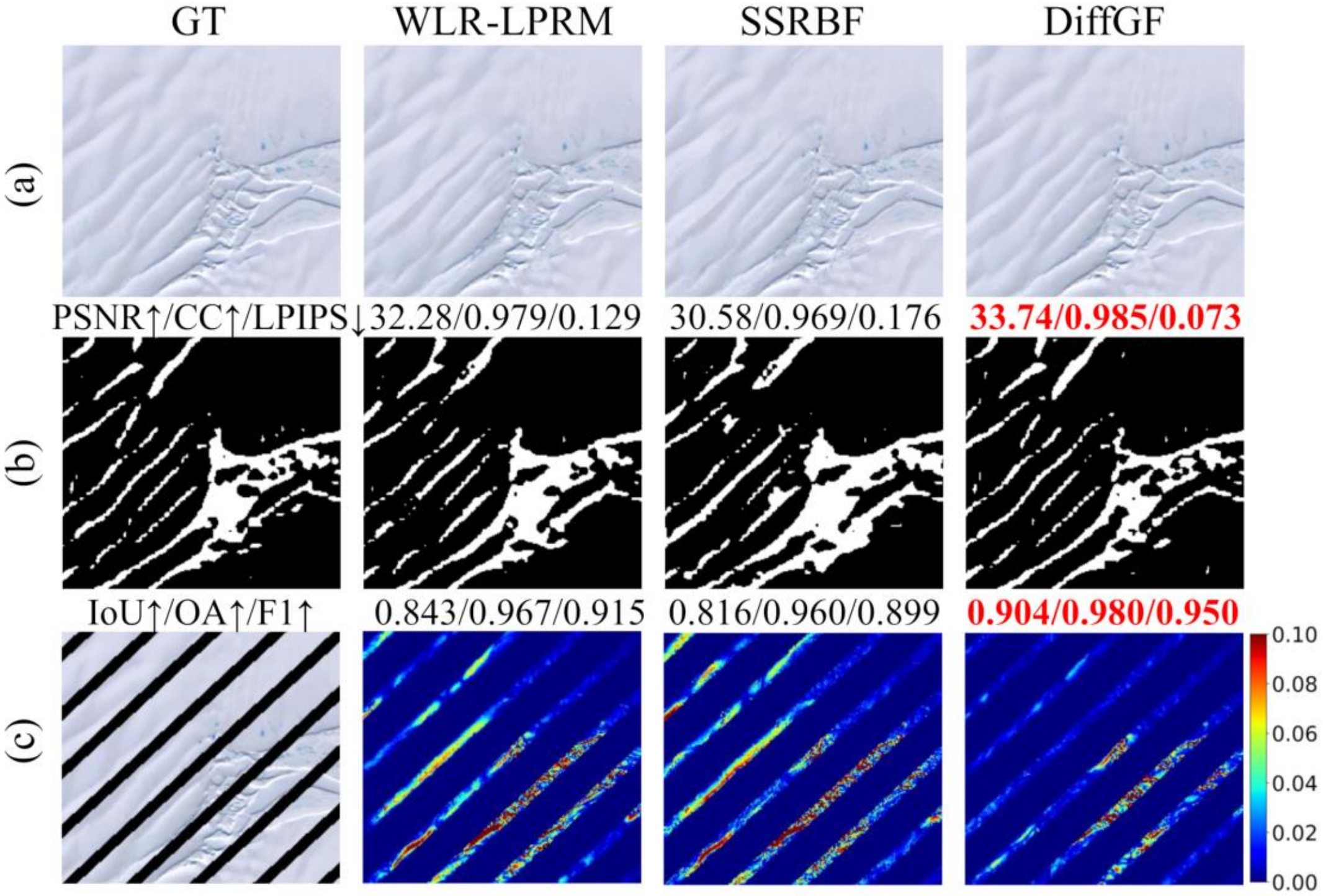

**Fig. 10.** Zoomed-in views of the red-boxed region in Fig. A7, showing detailed reconstruction results (a), crevasse segmentation results (b), and the SLC-off input together with corresponding error maps of the reconstructed images (c). The annotated metrics are computed over the entire image shown in Fig. A7.

SLC-off images in order to obtain complete ground truth. Although pre-failure Landsat 7 images could also be used for simulation, their time span is limited, especially in Antarctica, where usable optical imagery is restricted to the austral summer. Landsat 8 imagery was therefore selected to provide sufficient data for dataset construction. The RGB bands of Landsat 7 and 8 have slight wavelength differences, but they both correspond to the visible red, green, and blue bands, with the maximum difference between the corresponding band boundaries being approximately 0.02 μm. Since the restoration task in this study mainly focuses on spatial structure, texture continuity, and local spectral consistency rather than quantitative reflectance analysis, these slight differences are not expected to substantially affect the restoration results. Restoration experiments on real Landsat 7 SLC-off images, as shown in Figs. A2-A5, further support the applicability of DiffGF to real SLC-off image restoration.

### C. Motivation for Non-Reference Restoration in Antarctica

To further support the motivation for a non-reference restoration approach, we conduct an additional experiment to test the performance of reference-based methods in Antarctic applications. The simulated SLC-off image was acquired on December 21, 2021, and the co-located Landsat 8 reference image was acquired on October 2, 2021, with a temporal interval of 80 days. Due to the limited cloud-free overlap of the reference image, a smaller area of approximately 38.4 km × 53.8 km is used in this experiment (Fig. A7).

Two representative reference-based methods, WLR-LPRM [15] and SSRBF [16], are evaluated using their official implementation settings. The restored images are further used for downstream crevasse segmentation. The segmentation metrics, computed between the segmentation results derived from the restored images and those derived from the original complete image, are annotated in Fig. 10 following the same calculation method as in Table III. As shown in Fig. 10, the reference-based methods produce less favorable restoration and downstream crevasse segmentation results in this case, whereas DiffGF produces more consistent results. This example indicates that even a relatively short reference time interval can introduce challenges over rapidly changing Antarctic ice shelves, thereby providing experimental support for the motivation of developing a non-reference SLC-off restoration framework for Antarctic applications. It should be noted that SSRBF was originally designed using six bands, while only RGB bands are used here. Therefore, its result should be interpreted as a comparison under the data settings of this study rather than as its optimal performance.

### D. Limitations

Although DiffGF achieves competitive overall performance, its generative nature should be considered when interpreting the restored results. Accordingly, the restored content should be regarded as a plausible reconstruction of the missing regions. Its reliability may also vary depending on the quality of the input imagery.

When applying DiffGF to real Landsat 7 SLC-off imagery, we observed that the framework may fail for images affected by image quality issues, such as abnormal brightness or color distribution (Fig. A6). In such cases, the restoration within the SLC-off gaps lacks spectral and spatial continuity with the surrounding known regions. By contrast, imagery of normal quality acquired over the same region can be restored more effectively (Fig. A5). This limitation is likely because the SLCANT dataset used for training consists of images with normal visual and radiometric quality, and therefore does not sufficiently represent such degraded inputs.

## V. Conclusion

In this article, we propose DiffGF, a non-reference diffusion-based framework for restoring Landsat 7 ETM+ SLC-off imagery in Antarctica. DiffGF adopts a two-stage architecture, consisting of a latent-space diffusion process and a pixel-space refinement, termed the Mask-Guided Harmonization Network (MGHNet). By leveraging the generative capability of diffusion models and introducing a refinement stage, DiffGF reconstructs missing regions without relying on any external reference data.

To support training and evaluation, we construct a dedicated Antarctic dataset, SLCANT, which covers diverse surface characteristics and incorporates realistic SLC-off stripe patterns. Both quantitative and qualitative evaluations demonstrate that DiffGF achieves high-fidelity reconstruction. Moreover, a downstream crevasse segmentation task further suggests that DiffGF can restore visual appearance while preserving application-relevant semantic information.

Comparative experiments against representative SLC-off restoration and image inpainting methods show that DiffGF achieves competitive performance in both image quality metrics and application-oriented evaluations. These results indicate that DiffGF provides a useful approach for exploiting Landsat 7 SLC-off archives in Antarctica, enabling the extraction of valuable information from historical observations and supporting related Antarctic studies.

In future work, we will explore the extension of the DiffGF framework to other remote sensing restoration tasks, such as cloud removal, to examine its potential applicability to a broader range of remote sensing applications.

## Appendix

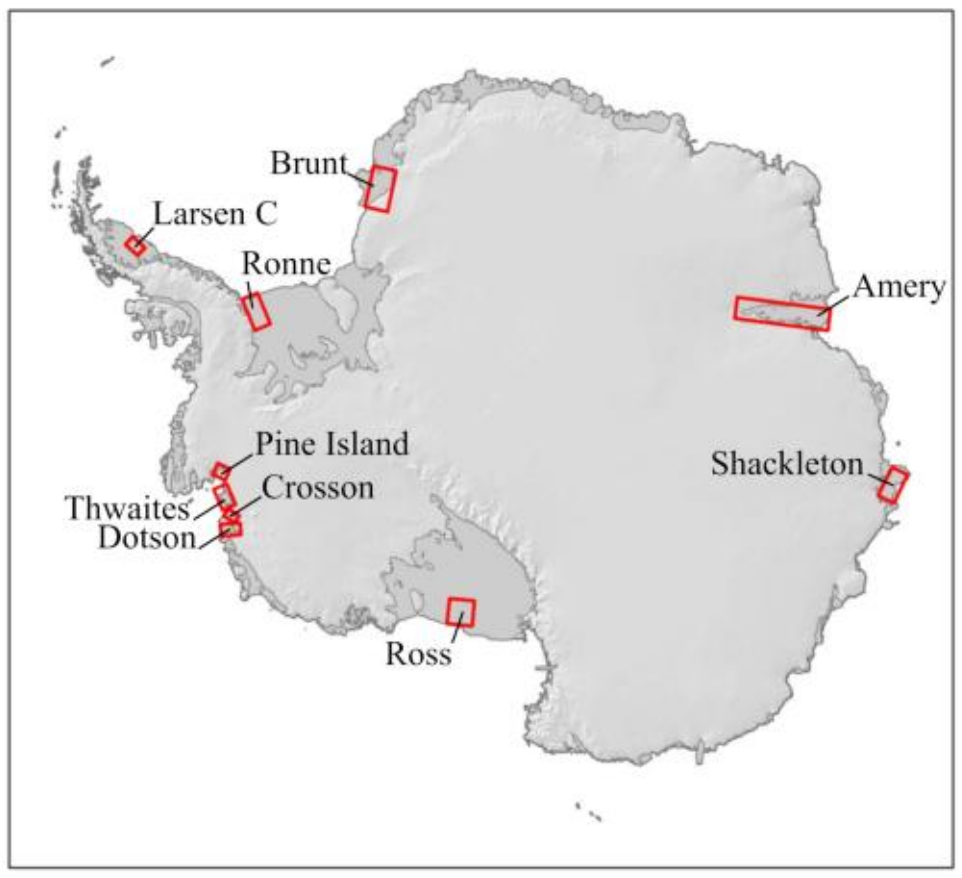

**Fig. A1.** Spatial distribution of the source imagery used for the construction of the SLCANT dataset. Red polygons indicate the extent of the selected imagery, with ice shelf names labelled. The background imagery and boundary data are from Bedmap2 [62].

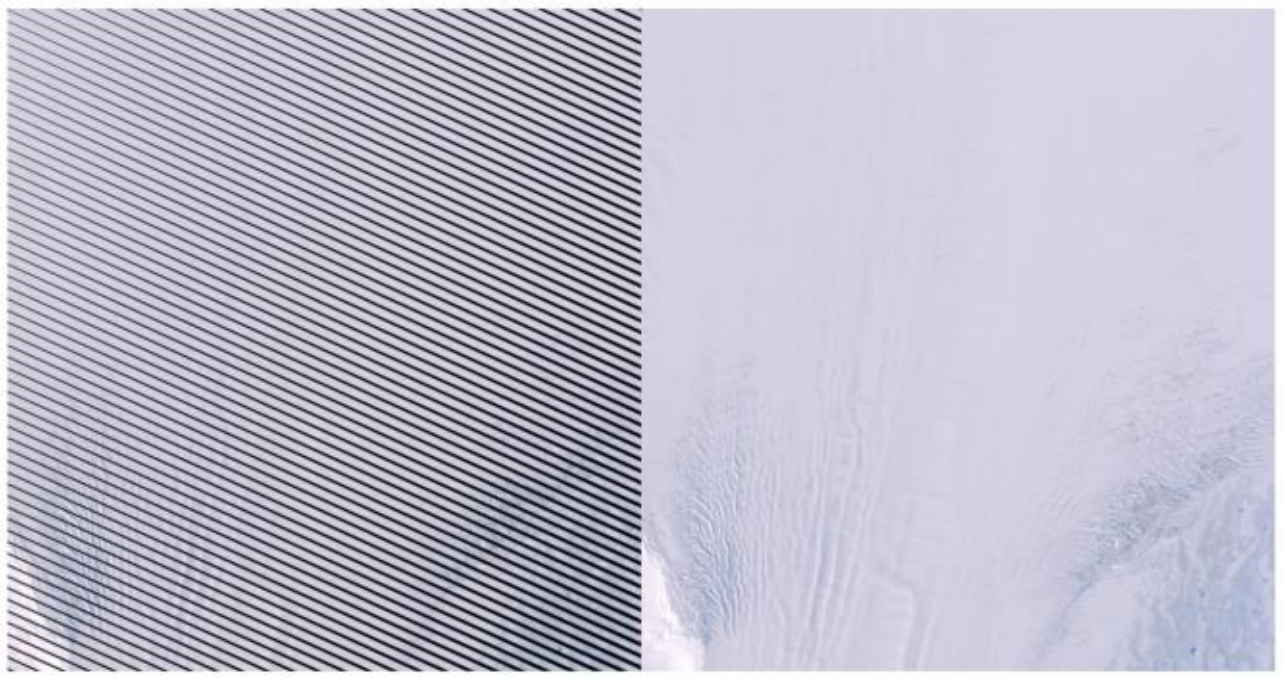

**Fig. A2.** Landsat 7 SLC-off imagery acquired over the Fimbul Ice Shelf on February 2, 2007 (left) and the corresponding DiffGF restoration result (right).

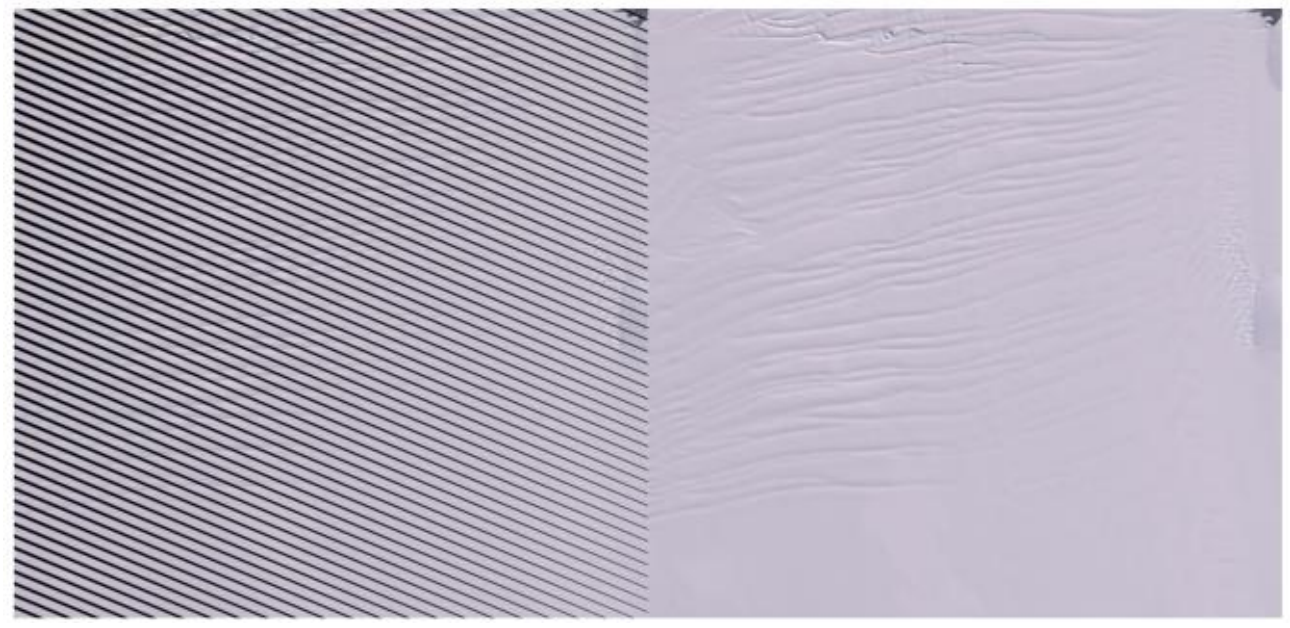

**Fig. A3.** Landsat 7 SLC-off imagery acquired over the Jelbart Ice Shelf on February 25, 2007 (left) and the corresponding DiffGF restoration result (right).

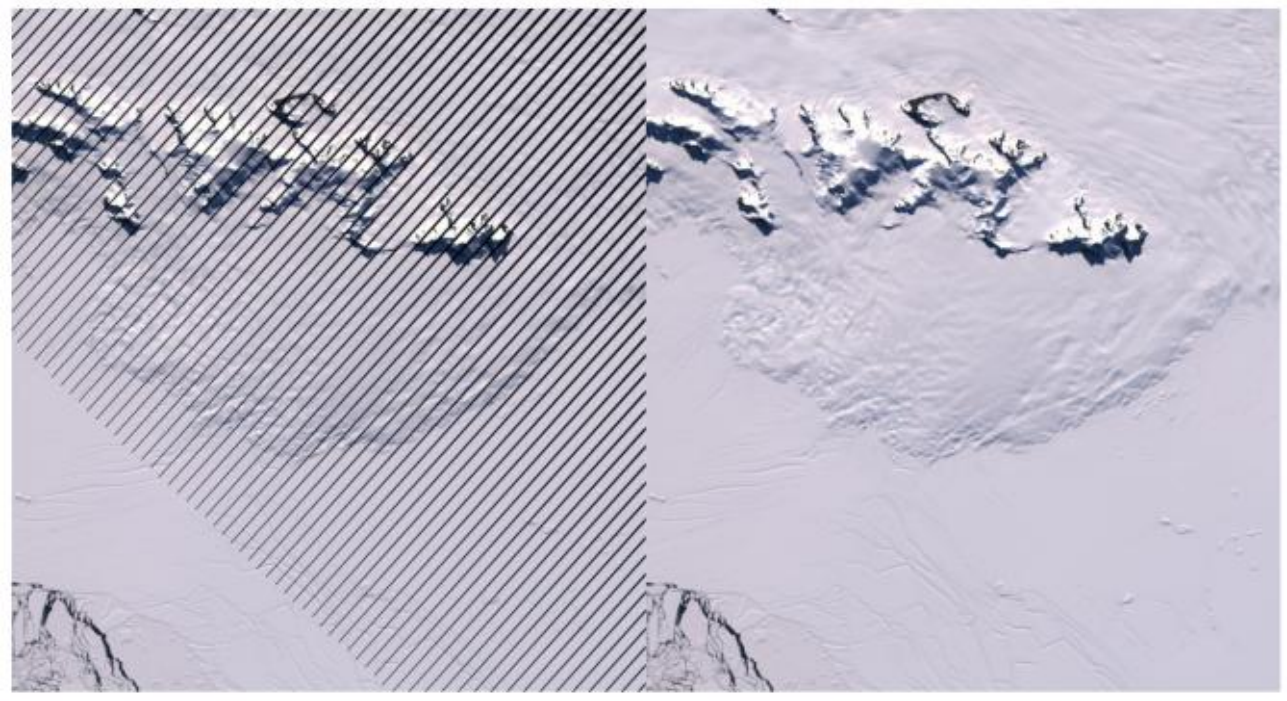

**Fig. A4.** Landsat 7 SLC-off imagery acquired over the Wilkins Ice Shelf on February 6, 2011 (left) and the corresponding DiffGF restoration result (right).

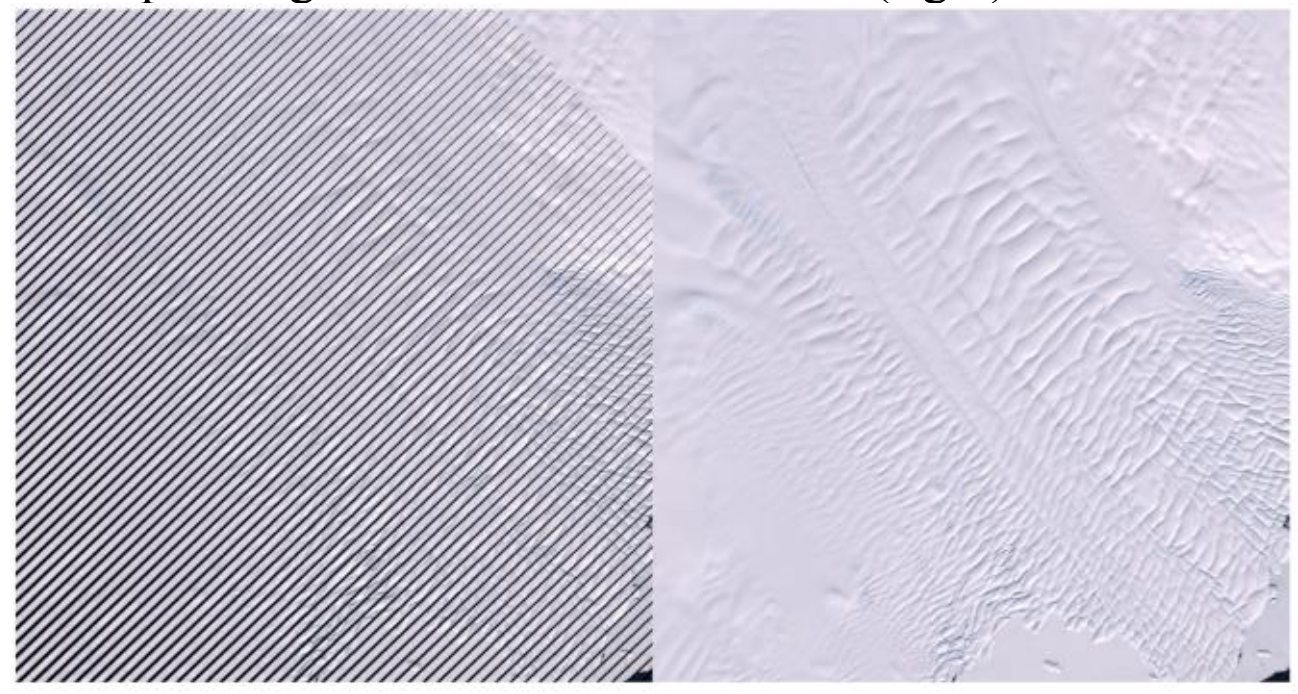

**Fig. A5.** Landsat 7 SLC-off imagery acquired over the Totten Ice Shelf on February 18, 2010 (left) and the corresponding DiffGF restoration result (right).

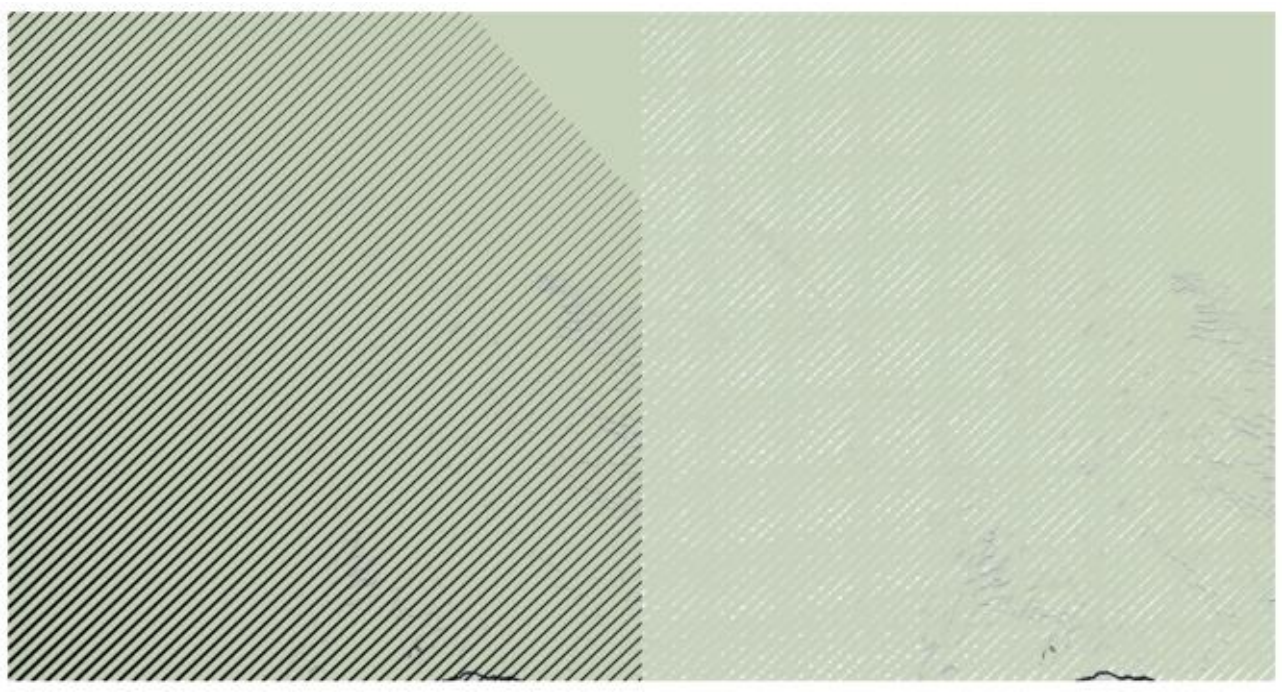

**Fig. A6.** Example of Landsat 7 SLC-off imagery affected by image quality issues, acquired over the Totten Ice Shelf on December 29, 2008: original SLC-off imagery (left) and the corresponding DiffGF restoration result (right).

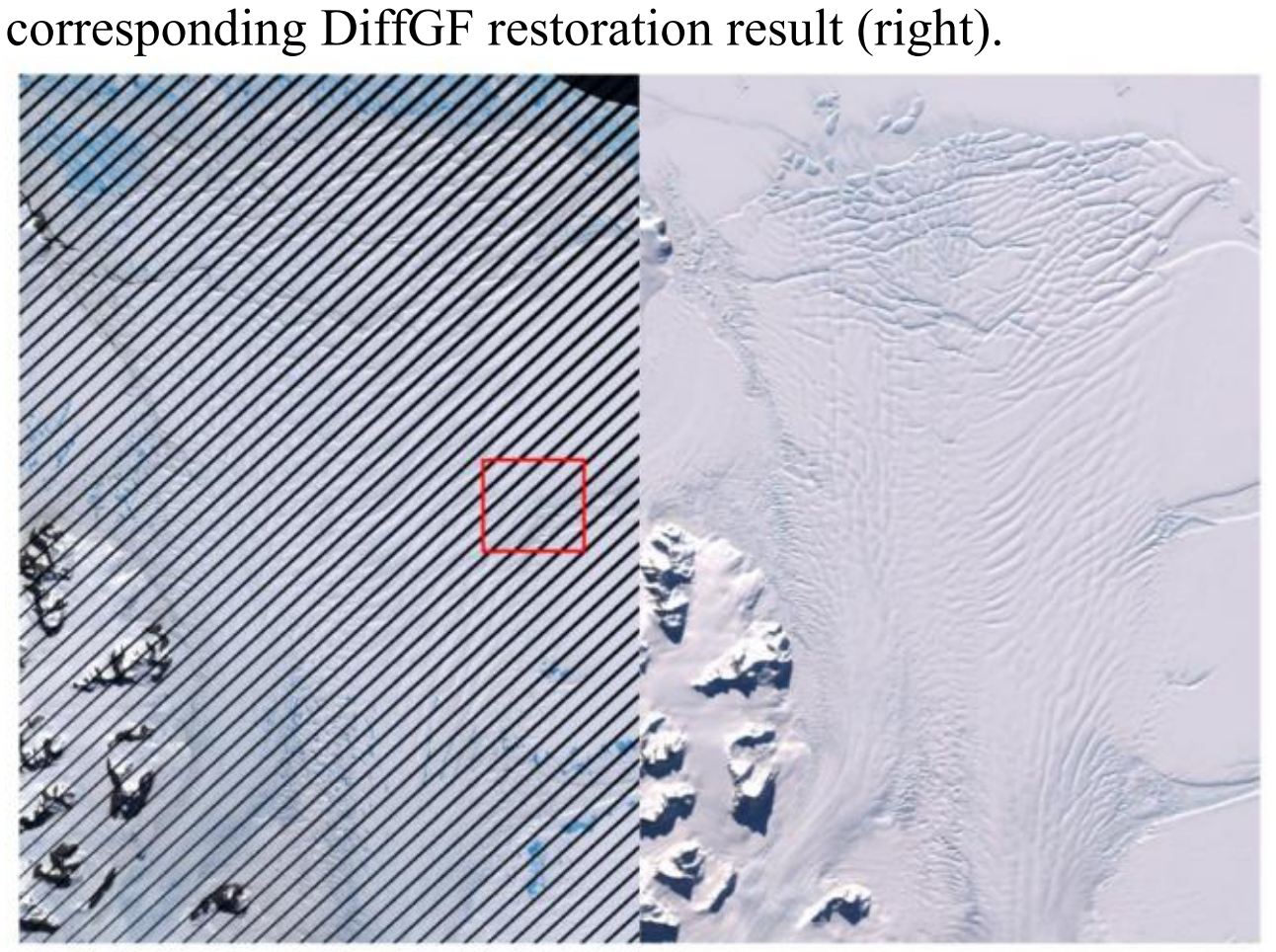

**Fig. A7.** Simulated SLC-off imagery generated from a Landsat 8 image acquired on December 21, 2021 (left), and the reference Landsat 8 image acquired on October 2, 2021 (right). The red box indicates the region shown in Fig. 10.

TABLE A1
SUMMARY OF SOURCE IMAGERY USED IN THE CONSTRUCTION OF THE SLCANT DATASET (DATES IN YYYY/MM/DD FORMAT).

| Ice Shelf/Glacier | Landsat 8 date | Landsat 7 date |
|---|---|---|
| Amery | 2020/01/20 | 2011/12/21 |
| Brunt | 2021/01/07 | 2012/01/16 |
| Crosson | 2019/01/10 | 2011/12/30 |
| Dotson | 2019/01/22 | 2013/02/05 |
| Larsen C | 2020/12/29 | 2012/12/31 |
| Pine Island | 2018/02/28 | 2013/02/01 |
| Ronne | 2021/01/15 | 2012/12/30 |
| Ross | 2021/01/24 | 2013/01/08 |
| Shackleton | 2019/02/16 | 2012/12/14 |
| | 2021/01/11 | 2012/12/12 |
| Thwaites | 2019/01/12 | 2013/01/10 |

## ACKNOWLEDGMENT

This work was carried out using the computational facilities of the Advanced Computing Research Centre, University of Bristol - http://www.bristol.ac.uk/acrc/.

During the preparation of this manuscript, ChatGPT was used in order to enhance the readability. After using this tool, the authors reviewed and edited the content as needed and take full responsibility for the content of the publication.